\documentclass[11pt]{article}
\usepackage[utf8]{inputenc}
\usepackage[margin=1in]{geometry}
\usepackage[english]{babel}

\usepackage{amsmath,amsthm,amssymb,amsfonts}
\usepackage{booktabs} 
\usepackage{tabularx}  
\usepackage{ragged2e}
\usepackage{authblk}

\usepackage{algorithmic}
\usepackage{algorithm}
\usepackage{array}
\usepackage[caption=false,font=normalsize,labelfont=sf,textfont=sf]{subfig}
\usepackage{textcomp}
\usepackage{stfloats}
\usepackage{url}
\usepackage{verbatim}
\usepackage{graphicx}
\usepackage{cite}
\usepackage[colorlinks=true, allcolors=blue]{hyperref}
\newtheorem{lem}{Lemma}[section]

\newtheorem{prop}{Proposition}[section]
\newtheorem{remark}{Remark}[section]
\newtheorem{theorem}{Theorem}[section]

\usepackage{booktabs}
\usepackage{multirow}
\usepackage{makecell}
\usepackage[table]{xcolor}
\usepackage[normalem]{ulem}  

\newcommand{\vtag}[2]{\rotatebox{90}{\makecell{\footnotesize#1\\ \footnotesize #2}}}

\newcommand{\rmd}{\mathrm{d}}
\newcommand{\T}[1]{\textbf{#1}}
\newcommand{\R}{\mathbb{R}}
\newcommand{\E}{\mathbb{E}}
\DeclareMathOperator*{\argmax}{arg\,max}

\newcommand{\Wtwo}{\mathcal{W}_2}

\title{COFM: Consistent Optimal Transport Flow Matching via Partially Input Convex Neural Networks}
\author[1]{Fanghui Song}
\author[2]{Zhongjian Wang\thanks{Corresponding author: zhongjian.wang@ntu.edu.sg}}
\author[1]{Jiebao Sun}

\affil[1]{School of Mathematics, Harbin Institute of Technology, 150006, Harbin, China}
\affil[2]{School of Physical and Mathematical Sciences, Nanyang Technological University, 637371, Singapore}

\begin{document}

\maketitle
\footnotetext{This work was supported by the National Natural Science Foundation of China under Grants 12171123 and 12271130, the NTU SUG-023162-00001, the MOE AcRF Tier 1 Grant RG17/24, and the Jing-Jin-Ji Regional Integrated Environmental Improvement-National Science and Technology Major Project (2026ZD1211102).  Part of the work by FS is done during her visit to NTU. The source code is available at \href{https://github.com/sfh623/COFM-Consistency-Optimal-Transport-Flow-Matching}{https://github.com/sfh623/COFM-Consistency-Optimal-Transport-Flow-Matching}.}

\maketitle

\begin{abstract}
Optimal transport (OT) provides a principled framework for learning mappings between probability distributions, and has found broad applications in generative modeling, inverse problems and scientific computing. Recently, flow matching methods have emerged as an efficient paradigm for learning continuous-time transport dynamics. However, existing OT-based flow matching methods often suffer from either high computational cost due to inner optimization or limited consistency. Moreover, it remains challenging to design neural architectures that can simultaneously guarantee convexity, stability, and efficient transport learning.
In this paper, we propose a framework for consistent optimal transport flow matching. Specifically, we parameterize the transport potential using partially input convex neural networks (PICNN), and incorporate a Hamilton–Jacobi residual into the training objective to enforce dynamical consistency of the learned flow. This design enables a unified formulation that supports both one-step transport andmulti-step ODE-based sampling, without requiring costly inner optimization.
Extensive experiments on benchmark datasets demonstrate that the proposed method achieves competitive performance compared with existing OT-based and flow matching approaches, while maintaining favorable computational efficiency. In particular, under the $D=256$ benchmark, COFM achieves more than a $2\times$ reduction in $\mathcal{L}^2$-UVP compared with state-of-the-art (SOTA) models, while requiring approximately $9\times$ less computational time. These results suggest that combining convex potential structures with HJ-based dynamical regularization provides an effective framework for scalable and geometrically consistent transport learning.
\end{abstract}


\section{Introduction}
Optimal
transport (OT) provides a principled framework for comparing and transforming probability distributions with minimal transport cost, and has found broad applications in generative modeling, distribution translation, image-to-image generation, and high-dimensional scientific data analysis  \cite{wang2022deepparticle,fan2023neural,li2025dpot}.

Classical numerical approaches discretize the domain and solve OT via the Monge-Amp\`{e}re PDE on grids, for example, using consistent finite-difference schemes with transport boundary conditions \cite{froese2011convergent,froese2012numerical}. While accurate and mathematically well--founded, such solvers suffer from the curse of dimensionality and become impractical in high dimensions or for complex support sets, thereby limiting their applicability to modern data-driven tasks.
To overcome these limitations, \T{network--based OT methods} parameterize Kantorovich dual potentials with neural networks and optimize the dual objective. Representative approaches include min–max training of neural OT potentials and amortized convex conjugate formulations that learn Wasserstein-2 ($\Wtwo$) maps \cite{korotin2021bm, makkuva2020optimal, korotin2021wasserstein}. Although these methods scale beyond grids and can yield high-quality maps, the saddle-point landscape and the need to enforce Lipschitz or convexity constraints often lead to delicate optimization dynamics and slower convergence at scale.

\T{Flow matching models (FMs)} learn a time-dependent vector field whose integral curves transport a simple base distribution to the data distribution, where training regresses the vector field toward prescribed target velocities at intermediate times, typically yielding more stable optimization than adversarial objectives. This perspective fits within a broader paradigm that models data generation as a continuous flow governed by an ODE. Early \T{normalizing flows} construct bijective neural transformations to enable exact density estimation \cite{rezende2015variational,lipmanflow,papamakarios2021normalizing}. More recently, \T{neural ODEs} reinterpret neural networks as continuous-time dynamical systems, providing a flexible and invertible framework for modeling flows \cite{neuralODE}. Crucially, the geometry of trajectories is learned rather than fixed: depending on the parameterization, conditioning, and boundary conditions, ODE paths can be nearly straight or markedly curved. Deterministic sampling works in both settings, although straighter paths generally reduce numerical stiffness and the number of function evaluations (NFEs).

Over the past decade, \T{diffusion models} have attracted significant attention. These methods learn denoisers or score fields along a noise schedule and generate samples by integrating a reverse-time SDE, with the probability-flow ODE as its deterministic counterpart \cite{song2020score,yang2023diffusion,gushchin2024light}. Despite delivering state-of-the-art image quality, diffusion samplers typically require tens or hundreds of NFEs and traverse highly curved trajectories in the state space. Most recently, MeanFlow \cite{meanflow} proposed regressing the average velocity field during training, thereby enabling high-quality one-step generation and accelerating inference in FM. While effective, regressing a mean velocity can average out mode-specific transport and fine geometric structures. More generally, standard diffusion training relies on heat flow, which neither minimizes transport cost nor encourages alignment with displacement interpolation.

Motivated by the efficiency and interpretability of the sample, recent work aims to \T{straighten flow trajectories via OT}. 
Rectified flow iteratively steers the learned ODE toward linear paths \cite{liu2022rectified}, while consistency flow matching imposes velocity consistency across different time origins \cite{yang2024consistency}. OT-Flow regularizes continuous normalizing flows with OT-inspired priors to encourage easy-to-integrate trajectories \cite{onken2021ot}.
While these methods improve sampling efficiency and quality, their training is typically conducted in discrete time. Moreover, the learned vector fields generally lack explicit constraints related to optimal transport geometry and temporal consistency, which may limit the structural coherence and interpretability of the resulting dynamics\cite{gat2024discrete}.

Building on these efforts,  {one-step flow-based OT seeks to learn vector fields whose trajectories approximate the displacement interpolation induced by optimal transport \cite{mccann1997convexity}}, allowing high-quality samples to be generated in a single integration step. \T{Optimal flow matching (OFM)} advances this idea by learning vector fields aligned with straight OT trajectories in one step \cite{OFM,tong2024improving}. However, many existing variants require solving auxiliary optimization subproblems during training, which introduces non-trivial computational overhead. 

Despite the recent progress, existing optimal transport-based FMs still suffer from several limitations. On the one hand, many approaches rely on costly inner optimization procedures, which significantly increases computational complexity. On the other hand, there is often a lack of consistency between one-step transport and multi-step sampling, leading to degraded performance under low-step inference settings.

To address the limitations identified above, we propose a \T{Consistent Optimal Transport Flow Matching (COFM)} framework that directly learns a time--dependent convex potential $\Psi(t,x)$ using a single partially input convex neural network (PICNN) \cite{ICNN2017}. 
The induced vector field is conservative by construction and globally structured by convexity, which is consistent with the convex structure arising in the Kantorovich dual formulation of quadratic optimal transport.
Beyond potential-based parameterization, we introduce an additional term in the loss function inspired by the Hamilton-Jacobi (HJ) formulation of displacement interpolation that enforces the temporal consistency of $\Psi$ and encourages evolution along displacement interpolation \cite{villani2009optimal,HJPINN}.

Unlike OFM-like methods, which require inner optimization subproblems, the COFM computes velocities by differentiating a single PICNN, thereby avoiding the expense of nested optimization. This yields a unified training that scales to high dimensions and supports principled one-step (or multi-step) transport with improved stability, interpretability, and sampling efficiency. 

 In particular, under $D=256$ benchmark, COFM achieves more than  a $2\times$ reduction in $\mathcal{L}^2$-UVP compared with OFM, while requiring approximately $9\times$ less computational time. Diffusion models typically require many function evaluations due to iterative denoising, whereas COFM supports stable one-step transport and multi-step sampling within a unified OT-based framework. Additionally, by explicitly parameterizing a convex potential aligned with optimal transport geometry, COFM encourages trajectories that are more consistent with displacement interpolation, which improves interpretability. 

Beyond its theoretical formulation, COFM is motivated by practical considerations in OT-based generative modeling. In many high-dimensional transport and generation tasks, it is desirable to obtain transport trajectories that remain geometrically structured and approximately minimal under the transport cost, while maintaining low computational cost during inference. However, existing OT-based approaches often rely on expensive inner optimization or require many sampling steps, which may limit their scalability and efficiency.

In contrast, the proposed COFM learns a time-dependent convex potential within a unified framework, enabling both stable one-step transport and multi-step sampling without additional inner optimization. This makes the method particularly suitable for distribution translation and latent-space image-to-image generation tasks, where preserving structured and geometrically consistent transport between source and target samples is important.

In summary, our main advantages are
\begin{itemize}
    \item We propose a convex potential conditioned by time parameterized by a PICNN, {that provides a principled and globally structured representation consistent with the optimal transport geometry, leading to improved training stability and model interpretability.}
    \item We introduce a novel regularization inspired by the HJ semigroup {to enforce temporal consistency of the learned dynamics and promote evolution along displacement interpolation, allowing one-step and multi-step sampling within a unified framework.}
    \item Our approach bypasses costly inner min-max optimization and auxiliary optimization subproblems, {resulting in a more efficient and scalable training procedure, especially in high-dimensional settings.}
\end{itemize}

The paper is organized as follows: Section \ref{sec2} provides background on related flow models and optimal transport, Section \ref{sec3} describes our proposed method in detail, Section \ref{sec5} presents extensive experimental results, and Section \ref{sec6} provides conclusions and limitations.

\section{Preliminaries and related work}\label{sec2} 
\subsection{Optimal Transport and Hamilton--Jacobi Background}
 {We briefly review the OT and HJ preliminaries relevant to our framework, and refer to \cite{villani2009optimal} for a comprehensive treatment.}

 Let $\mu_0,\mu_1\in\mathcal{P}_2(\mathbb{R}^d)$ be two probability
measures with finite second moments.
Under the quadratic cost
\begin{equation}\label{eq:costfunc}
  c(x,y)=\tfrac12\|x-y\|^2,
\end{equation}
the optimal transport problem admits a unique solution in the form of a
gradient map, known as the \emph{Brenier map},
\begin{equation}\label{eq:brenier}
  T(x)=\nabla \Psi^\ast(x),
\end{equation}
where $\Psi^\ast$ is a convex potential.

 The corresponding squared $2$-Wasserstein distance is
\begin{equation}\label{eq:w2def}
  \Wtwo^{2}(\mu_{0}, \mu_{1})
  = \min_{\pi\in\Pi(\mu_0,\mu_1)}
  \int \tfrac12\|x_{1}-x_{0}\|^{2}\,\mathrm d\pi(x_0,x_1).
\end{equation}

Equivalently, the quadratic OT problem admits a single-potential dual formulation. 
For a convex function $\Psi:\mathbb{R}^d\to\mathbb{R}$, let
\[
  \bar\Psi(x_1):=\sup_{x_0\in\mathbb{R}^d}
  \big\{\langle x_0,x_1\rangle-\Psi(x_0)\big\}
\]
be its convex conjugate. Then the OT objective can be written as
\begin{equation}\label{eq:w2dual}
    \mathcal{L}_{\mathrm{OT}}(\Psi)
    := \int_{\mathbb{R}^d}\Psi\,\mathrm d\mu_0
      + \int_{\mathbb{R}^d}\bar\Psi\,\mathrm d\mu_1
\end{equation}
with $\Psi$ constrained to be convex. This dual form underlies potential-based OT methods and motivates the use of convex neural parameterizations.

 In practical learning-based OT methods, the transport coupling is often approximated at the minibatch level by solving a discrete OT matching problem between sampled particles. For example, OT-CFM \cite{tong2024improving} constructs batch-wise pairings by solving a discrete OT problem with quadratic cost
\[
C_{ij}=\tfrac12\|z_i-x_j\|^2,
\]
where the resulting coupling is used to define supervised transport trajectories during training.

A fundamental property of OT is the \emph{displacement interpolation}.
Given the optimal map $T$, define
\[
T_t(x)=(1-t)x+tT(x), \qquad \mu_t=(T_t)_\#\mu_0.
\]
Then $(\mu_t)_{t\in[0,1]}$ forms a constant-speed geodesic in
$\Wtwo$, where each particle moves along a straight line.

This displacement structure is closely related to the
Hamilton--Jacobi (HJ) equation. In particular, the Hopf--Lax operator
generates a solution to
\begin{equation}\label{eq:HJequation}
  \partial_t u + \tfrac12 \|\nabla u\|^2 = 0,
\end{equation}
whose characteristics correspond to straight-line trajectories with
velocity determined by the gradient of a potential. This connection
provides a dynamic interpretation of OT and motivates our use of an
HJ-type residual to regularize the learned flow. 

In addition, physics-informed machine learning methods, such as physics-informed neural networks (PINNs) \cite{raissi2019physics, sirignano2018dgm, e2018deep, han2018solving,ROY2023472}, incorporate governing equations into neural networks via residual-based constraints, and have been widely applied to modeling various physical and constitutive systems, as well as solving high-dimensional PDEs and control equations \cite{han2018solving,nakamura2021adaptive}.


Recent studies further extend this paradigm to complex engineering and physical phenomena \cite{ROY2023106049, ROY2024105570, BOSE2024107483,roy2024combining}, demonstrating the effectiveness of residual-based learning in capturing underlying dynamics. While these approaches are primarily designed to approximate solutions of prescribed PDEs, our framework differs in that the Hamilton--Jacobi residual is not used to solve a standalone HJ equation. In particular, the corresponding HJ structure in our setting is not given as a fully prescribed boundary value problem with known transport coupling and geometry, and therefore does not constitute a standard well-posed PDE system that can be directly solved using conventional PINN formulations. Instead, the proposed residual serves as a soft regularization mechanism within an optimal transport and flow matching framework to encourage temporal consistency and displacement-interpolation-like transport behavior.

\subsection{Flow matching models}

Consider modeling data generation by a continuous-time flow governed by the ODE \cite{liu2022rectified}
\begin{equation}\label{eq:ode}
  \dot X_t \;=\; v_\theta(X_t,t),
  \qquad
  t\in[0,1],
  \qquad
  X_0\sim p_{\mathrm{init}},
\end{equation}
where $v_\theta:\R^d\times[0,1]\to\R^d$ is a time-dependent neural velocity field, and $p_{\mathrm{init}}$ denotes a prescribed source distribution.

Let $\psi^{\theta}_{s\to t}:\R^d\to\R^d$ denote the flow map associated with \eqref{eq:ode}, which transports a point from time $s$ to $t$. In particular,
\[
X_t=\psi^{\theta}_{0\to t}(X_0),
\]
and the induced time-dependent distribution $p_t$ satisfies
\begin{equation}\label{eq:push}
  (\psi^{\theta}_{0\to t})_{\#}p_{\mathrm{init}}
  \;=\;
  p_t.
\end{equation}

Under suitable regularity assumptions, the density evolution of $\{p_t\}_{t\in[0,1]}$ is governed by the continuity equation
\begin{equation}\label{eq:continuity}
  \partial_t p_t
  +
  \nabla\!\cdot\!\big(p_t\, v_\theta(\cdot,t)\big)
  =0,
  \qquad
  p_0=p_{\mathrm{init}}.
\end{equation}
The objective is to learn $v_\theta$ such that the terminal distribution satisfies
\(
p_1 \approx p_{\mathrm{data}}.\)
Samples are generated by first drawing $X_0\sim p_{\mathrm{init}}$ and then integrating \eqref{eq:ode} from $t=0$ to $1$ to obtain $X_1$.

FM trains $v_\theta$ by regressing it to a \emph{target instantaneous velocity} defined on a user-chosen \emph{bridging process} between $p_{\mathrm{init}}$ and $p_{\mathrm{data}}$.
A common choice is the straight-line path
\begin{equation}\label{eq:bridge}
  x_t \;=\; (1-t)\,x_0 + t\,x_1,\, (x_0,x_1)\sim \pi,\, t\sim\mathrm{Unif}[0,1],
\end{equation}
where $\pi$ is some coupling between $p_{\mathrm{init}}$ and $p_{\mathrm{data}}$ (e.g.\ the independent product $p_{\mathrm{init}}\!\times\!p_{\mathrm{data}}$ or a minibatch OT coupling).
Along \eqref{eq:bridge}, the pathwise velocity is ${\dot x} _ {t} = x_1-x_0$.
Conditioning on the intermediate state $x_t$ yields the conditional target
\begin{equation}\label{eq:target}
  u^\star(x,t) \;=\; \E\big[\,x_1-x_0 \,\big|\, x_t{=}x,\, t\,\big],
\end{equation}
and the \emph{flow-matching loss}
\begin{equation}\label{eq:fm_loss}
  \mathcal{L}_{\mathrm{FM}}(\theta)
  \;=\; \E_{t,(x_0,x_1)}\,
  \E_{x\,|\,x_t}\Big[\, \big\| v_\theta(x,t) - u^\star(x,t) \big\|^2 \,\Big].
\end{equation}
Minimizing \eqref{eq:fm_loss} encourages $v_\theta$ to reproduce the Eulerian velocity of the bridging process, which in turn transports $p_{\mathrm{init}}$ toward $p_{\mathrm{data}}$ when integrating \eqref{eq:ode}.
In practice, taking $\pi=p_{\mathrm{init}}\times p_{\mathrm{data}}$ yields the simple pathwise velocity $x_1-x_0$, while using the conditional expectation \eqref{eq:target} reduces variance and improves statistical efficiency.

The learned \emph{endpoint map} is $T_\theta \!:=\! \psi^{\theta}_{0\to1}$.
When $v_\theta$ is parameterized as a conservative field
$v_\theta=\nabla\Psi_\theta(t,x)$,
the induced dynamics are governed by a time-dependent potential and are naturally related to Brenier-type transport maps in optimal transport.

\subsection{Optimal flow matching} In \cite{OFM}, the authors restrict FM to the class of vector fields induced by a convex potential, thereby aligning the learned dynamics with the structure of quadratic-cost OT. Let $\Psi: \mathbb{R}^d\to\mathbb{R}$ be convex and define
\begin{align*}
   & \phi_{t}^{\Psi}:=(1-t) I+t \nabla \Psi, \\& x_{t}:=\phi_{t}^{\Psi}\left(z_{0}\right)=(1-t) z_{0}+t \nabla \Psi\left(z_{0}\right), \quad t \in[0,1] .
\end{align*}
Here $z_0$ denotes the latent preimage of $x_t$ under $\phi_{t}^{\Psi}$, the unique $z_0$ solving $x_t=\phi_{t}^{\Psi}(z_0)$ equivalently,
\begin{equation}\label{ofm-ot}
    z_{0}=\arg \min _{z}\left[\frac{1-t}{2}\|z\|^{2}+t \Psi(z)-\left\langle x_{t}, z\right\rangle\right].
\end{equation}
As stated in \cite{OFM}, the optimization subproblem \eqref{ofm-ot} is $(1-t)$-strongly convex, while the closed form solution is unknown. Hence, the numerical solver involves computing the gradient of $\Psi$ through {\sc Autodiff}.   
The time-dependent vector field generated by $\Psi$ is 
\[u_{t}^{\Psi}\left(x_{t}\right):=\nabla \Psi\left(z_{0}\right)-z_{0}, \quad \text { where } z_{0}=\left(\phi_{t}^{\Psi}\right)^{-1}\left(x_{t}\right) .\]
Given a coupling $\pi\in \Pi(\mu_0,\mu_1)$, the displacement interpolation $x_t=(1-t)x_0+tx_1$ with constant speed $\dot{x}_{t}^{\pi} \equiv x_{1}-x_{0}$, then OFM loss is 
\[\mathcal{L}_{\mathrm{OFM}}^{\pi}(\Psi)=\int_{0}^{1}\mathbb{E}_{(x_0,x_1)\sim\pi}\left\|u_{t}^{\Psi}\left(x_{t}\right)-\left(x_{1}-x_{0}\right)\right\|^{2}\mathrm{d}t\]
        
\begin{prop}
 For any $x_0,x_1\in \R^d$ and a convex function $\Psi$, the following equality holds
 \begin{align}\label{ofmmain}
\int_{0}^{1}\left\|u_{t}^{\Psi}\left(x_{t}\right)-\left(x_{1}-x_{0}\right)\right\|^{2}\mathrm{d}t=2\left[\Psi\left(x_{0}\right)+\bar{\Psi}\left(x_{1}\right)-\left\langle x_{0}, x_{1}\right\rangle\right].
 \end{align}
\end{prop}
 Under the quadratic cost \eqref{eq:costfunc} and assuming $\mu_0$ is absolutely continuous, the minimizer of $\mathcal{L}_{\mathrm{OFM}}^{\pi}(\Psi)$ over convex $\Psi$ coincides with the minimizer of the OT dual \eqref{eq:w2dual}, the Brenier potential $\Psi^*$. Consequently, OFM is designed to approximate OT geodesics $x_{t}=(1-t) x_{0}+t \nabla \Psi\left(x_{0}\right)$. In practice, OFM achieves a markedly straighter sampling path and lower transport cost than unconstrained FM models. However, OFM commonly requires inner subproblems \eqref{ofm-ot} (e.g., partial inversion or convex solves) and differentiation through them, which increases training time, especially in high dimensions. 

\section{Method}\label{sec3}
\subsection{Problem setup and learning objective}\label{sec:obj}
We consider the time interval $[0,1]$ and use the rescaled quadratic dynamic cost $c^{s,t}(x,y)=\frac{1}{2(t-s)}\|x-y\|^2$ for all $0\le s<t\le 1$.
Taking $t=1$, the $c^{s,t}$–optimal destination reduces to the straight displacement $$y=x+(1-s)\nabla\psi(s,x),$$ which motivates the introduction of a time-dependent $c^{s,1}$-convex potential $\psi(s,\cdot)$ whose gradient fields steer the flow.
Correspondingly, we consider $\Psi:[0,1]\times\R^d\to\R$, which is defined by the rescaling
\begin{equation}
    \label{eq:scalepsi}\Psi(t,x)=\tfrac12\|x\|^2+(1-t)\,\psi(t,x).
\end{equation}
This rescaling converts the $c$-convexity of $\psi$ into the standard convexity of $x\mapsto\Psi(t,x)$ at every $t$ and anchors the terminal condition at $t=1$ via $\Psi(1,x)=\tfrac12\|x\|^2$. The corresponding velocity field at time $t$ is defined as the gradient of the unscaled potential,
\begin{align}\label{eq:vel}
v_t(x)=\nabla\psi(t,x)=\frac{\nabla\Psi(t,x)-x}{1-t},\qquad t<1.
\end{align}
Note that, it is not necessary to evaluate $v_t$ at $t=1$ in either training or generation. The endpoint $t=1$ is handled by the exact anchor $\Psi(1,x)=\tfrac12\|x\|^2$, and one-step mapping from time $t$ follows $T_t(x)=\nabla\Psi(t,x)$.

From now on, we use $\Psi_\theta$ to denote the approximation of $\Psi$ by neural networks (detailed in Section~\ref{sec:PICNN}) under the following training objectives.
\paragraph{Flow Matching Loss}
We adopt the linear interpolation \(X_t=(1-t)X_0+tX_1\) with \((X_0,X_1)\sim\pi\). The flow matching term fits a potential $\psi_\theta$ whose spatial gradient reproduces this target velocity, i.e., the FM loss is defined as
\begin{align*}
\mathcal{L}_{\mathrm{FM}}(\psi_\theta)
&:=\int_0^1 \mathbb{E}_{(X_0,X_1)\sim\pi}\Big[\big\|X_1-X_0-\nabla\psi_\theta(t,X_t)\big\|^2\Big]\,\mathrm{d}t.
\end{align*}
Equivalently, using the potential scaled in time $\Psi^\theta$ by \eqref{eq:scalepsi}, we implement
\begin{equation}
 \label{eq:FM-term} \mathcal{L}_{\mathrm{FM}}(\Psi^\theta) =\int_0^1 \mathbb{E}_{(X_0,X_1)\sim\pi}
\|e_{\mathrm{FM}}(t,X_t)\|^2\rmd t, \end{equation}
where 
\begin{equation}\label{def:efm}
    e_{\mathrm{FM}}(t,X_t):=(X_1-X_0)+\frac{X_t-\nabla\Psi^\theta(t,X_t)}{1-t}
\end{equation}

\paragraph{HJ residual loss}
To promote temporal coherence of the learned dynamics, we add a HJ residual, in the $\psi$-form, which encourages consistency with \eqref{eq:HJequation}
via
\begin{align*}
\mathcal{L}_{\mathrm{res}}(\psi_\theta)
:=&\int_0^1 \lambda(t)\,\mathbb{E}_{(X_0,X_1)\sim\pi}\\&\Big|
\partial_t\psi_\theta(t,X_t)+\frac12\|\nabla\psi_\theta(t,X_t)\|^2
\Big|^2\,\mathrm{d}t.
\end{align*}
Here, we require $\pi$ to be the optimal transport plan. 
In the rescaled $\Psi$-form used in our implementation, the equivalent weighted residual is
\begin{equation}
\begin{aligned} \label{eq:HJ-term} \mathcal{L}_{\mathrm{res}}(\Psi^\theta)=\int_0^1 \lambda(t)\,\mathbb{E}_{(X_0,X_1)\sim\pi}\|\mathcal{R}(t,X_t)\|^2\mathrm{d}t, 
\end{aligned} 
\end{equation}
where
\begin{align}
    \mathcal{R}(t,X_t):=& \partial_t\Psi^\theta(t,X_t) +\frac{1}{1-t}\Big[\frac12\|\nabla\Psi^\theta(t,X_t)\|^2\nonumber\\&-X_t\cdot\nabla\Psi^\theta(t,X_t) +\Psi^\theta(t,X_t)\Big]. \label{def:R}
\end{align}
We choose $\lambda(t)=(1-t)^2$ to stabilize the objective near the terminal time. 
\paragraph{Pushforward consistency loss} 
While the HJ residual is effective in low dimensions, explicitly evaluating
\(\partial_t\Psi^\theta\), \(\nabla\Psi^\theta\), and their relative magnitudes can be numerically fragile in high dimensions. To mitigate this, we introduce a \emph{path-based pushforward consistency loss} that enforces velocity consistency via a \emph{discrete characteristic update}, thereby avoiding explicit \(\partial_t\Psi^\theta\).

At each iteration, we first sample \(X_0\sim\mu_0\), then draw \(t\sim\mathrm{Unif}[0,1-\delta]\). Using the network velocity
$v_\theta=\nabla\psi^\theta,$
we form a \emph{discrete characteristic point} by one explicit Euler step
\[
\widetilde X_t \;=\; X_0 + t\,v_\theta(0,X_0).
\]
We compare the terminal point along the same characteristic in $X_0$ and $\tilde{X}_t$ via
\[
r_{\mathrm{pf}}(t)\;=\;\nabla\Psi^\theta\!\big(t,\widetilde X_t\big)\;-\;\nabla\Psi^\theta\!\big(0,X_0\big),
\]
and define the weighted loss
\begin{align*}
&\mathcal{L}_{\mathrm{pf}}(\Psi^\theta)\;=\;
\mathbb{E}\!\left[\int_{0}^{1-\delta} w_{\mathrm{pf}}(t)\,\big\|r_{\mathrm{pf}}(t)\big\|^{2}\,\mathrm{d}t\right],
\\
&w_{\mathrm{pf}}(t)=(1-t)^{k},\;\;k\ge 0.
\end{align*}
In practice, we choose $\delta=10^{-2}$ and set \(k=4\) to strongly downweight the neighborhood near \(t\to1\), thereby mitigating endpoint potential oscillation due to \eqref{eq:vel}. To be noted, this construction does \emph{not} require sampling from \(\mu_1\) or solving any optimization subproblem.

Furthermore, direct computation shows,
\begin{equation*}
 \frac{\rmd }{\rmd t} r_{\mathrm{pf}}(t)=  \frac{\rmd }{\rmd t}\nabla \Psi(t,X_t)=\frac{1}{1-t}\nabla\mathcal{R} (t,X_t),
\end{equation*}
indicating the equivalence between $\mathcal{L}_{\mathrm{res}}$ and $\mathcal{L}_{\mathrm{pf}}$.

\medskip
Overall, the training objective in our Consistent Optimal Transport Flow Matching (COFM) approach reads,  
\begin{align}\label{eq:wholeloss}
\mathcal{L}_{\mathrm{COFM}}(\Psi^\theta):=\mathcal{L}_{\mathrm{FM}}(\Psi^\theta)+\mathcal{L}_{\mathrm{HJ}}(\Psi^\theta), 
\end{align}
where $\mathcal{L}_{\mathrm{HJ}}$ takes either $\mathcal{L}_{\mathrm{res}}$ or $\mathcal{L}_{\mathrm{pf}}$.
\subsection{Time-dependent 
PICNN}\label{sec:PICNN}
Let the network output the rescaled HJ potential
\(\psi_\theta:[0,1)\times\mathbb{R}^d\to\mathbb{R}\),
and define the convex potential $\Psi^\theta(t,x)$ by the rescaling, analogous to \eqref{eq:scalepsi}. 
Under \eqref{eq:vel}, we evaluate losses for $t\in[0,1)$ (i.e., we do not query $t=1$ where $(1-t)^{-1}$ appears).

Then parameterize the potential with a time-dependent PICNN.
Let \(z^{(0)}=0\). For \(l=1,\dots,L\), define the pre-activation
\begin{equation}\label{eq:icnn-pre}
\mathrm{pre}^{(l)}(t,x)=W_x^{(l)}x+W_{z}^{(l)}z^{(l-1)}+b^{(l)}+S_l(t),
\end{equation}
where \(S_l(t)\) is an MLP that depends on $t$ only. We then apply channel-wise {\sc ActNorm} \cite{kingma2018glow} with strictly positive per-channel scales, followed by an elementwise convex and nondecreasing activation~$\phi$:
\[
z^{(l)}(t,x)=\phi\,\!\Big(\mathrm{ActNorm}_l(\mathrm{pre}^{(l)}(t,x))\Big),\quad l=1,\dots,L-1,
\]
and keep the last layer linear: \(z^{(L)}(t,x)=\mathrm{pre}^{(L)}(t,x)\).

\begin{figure*}
    \centering
\includegraphics[width=0.6\linewidth, trim=1.5cm 1.5cm 2.5cm 3cm, clip]{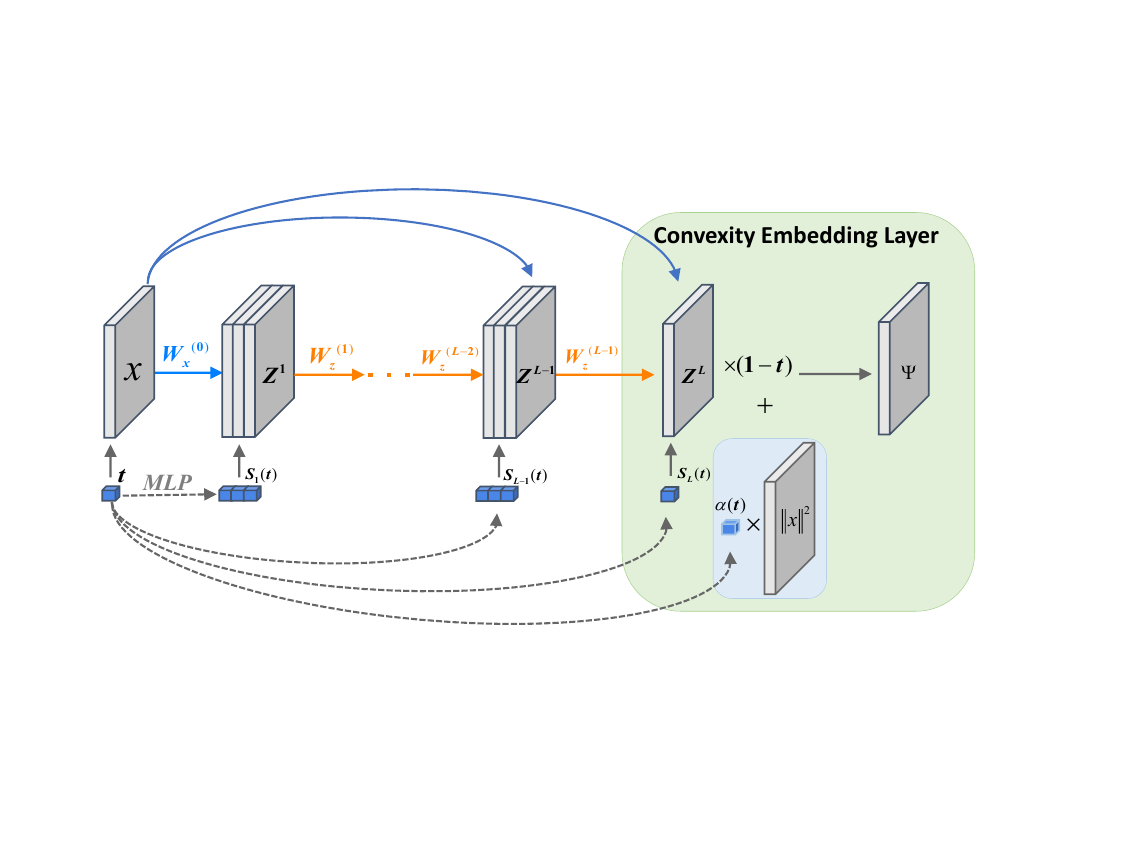} \vspace{-4em}
    \caption{\textbf{Time-dependent PICNN.} {Dashed paths denote values produced by MLPs (e.g., $S_l(t)$ and $\alpha(t)$),
while solid links represent affine transformations: blue for unconstrained
maps, orange for positive-weight maps, and black for identity connections.
The inputs are $(x,t)$.
Each layer receives an affine input mapping $W_x^{(l)}x$ and a
time-dependent bias $S_l(t)$, while $W_z^{(l)}$ propagates
$z^{(l-1)}$ forward under non-negativity constraints.
The final hidden output $z^{(L)}(t,x)$ is fed into the convexity embedding
layer (CEL), where it is gated by $(1-t)$ and combined with a
time-dependent quadratic term $\alpha(t)\|x\|^2$.
A single forward pass produces $\Psi^\theta(t,x)$, and the velocity field
is obtained via automatic differentiation with respect to $x$.}}
    \label{fig:ICNN}
\end{figure*}

The proposed PICNN satisfies the following constraints:
(i) \(W_{z}^{(l)}\ge 0\) elementwise;
(ii) \(\phi\) is convex and nondecreasing; 
(iii)  {we use the Softplus function as the activation in practice;}
(iv) \(S_l(t)\) depends only on $t$.
To enforce positivity of the transport weights, we parameterize
\[
W_{z}^{(l)}=\mathrm{Softplus}\big(\tilde W_{z}^{(l)}\big),
\]
so that positivity is guaranteed by construction. Under (i)–(iv), for fixed $t$, the mapping $x\mapsto z^{(L)}(t,x)$ is convex, and no positivity constraint is imposed on $W_x^{(l)}$ or $b^{(l)}$.

We generate $\alpha(t)$ from an MLP of $t$. 
For a positive and numerically stable time-gated quadratic, we use a factored
parameterization with an unconstrained scalar $r(t)=\mathrm{MLP}(t)$ and define
\begin{equation}\label{alpha_def}
\alpha(t)=\sigma\!\big(r(t)\,(1-t)\big),
\end{equation}
where $\sigma$ denotes the sigmoid function.
By construction,
\[
\alpha(1)=\sigma(0)=\tfrac12,\qquad
\alpha(t)\in(0,1)\ \text{for } t\in[0,1),
\]
hence
$\alpha(t)-\tfrac12=O(1-t)$.


We then define the Convexity Embedding Layer (CEL) as
\begin{equation}\label{eq:CELdef}
\Psi^{\theta}(t,x)=(1-t)\,z^{(L)}(t,x)+\alpha(t)\,\|x\|^{2}.
\end{equation}
Fig.~\ref{fig:ICNN} illustrates the time-dependent PICNN architecture.
Its main advantages are as follows.

\T{Convex potential and well-behaved endpoint.} With $W_z^{(l)}\geq 0$, a convex, nondecreasing activation $\phi$, and {\sc ActNorm} with strictly positive scales, the mapping $x\mapsto z^{(L)}(t,x)$ is convex for each $t$. Together with $\alpha(t)\!\ge0$, the CEL yields a convex $\Psi^\theta(t,\cdot)$ and therefore is consistent with OT theory.

In the temporal direction, per-layer time biases $S_l(t)$ provide fine-grained temporal modulation of convexity. 
 Moreover, in the CEL definition \eqref{eq:CELdef}, $\Psi^\theta(1,x)=\tfrac12\|x\|^2$ holds exactly, which improves conditioning over $t$ and avoids numerical sensitivity as $t\to1$.

\T{Efficiency and consistency.} 
The time-dependent PICNN produces a convex potential $\Psi^\theta(t,x)$ in one forward pass. Built on this, one can employ the auto-differentiation to compute the learning objective  $\mathcal{L}_{\mathrm{COFM}}$ defined in \eqref{eq:wholeloss}. 
Thus, each training update costs only one backward propagation through the same PICNN, without auxiliary or inner optimizers. 

At the inference stage, we can either (i) use the pushforward map $T_\theta(x)=\nabla\Psi^\theta(0,x)$ for \emph{one-step} sampling or (ii) solve the ODE \eqref{eq:ode} with $v_t=\nabla\psi_\theta$ for \emph{multi-steps} sampling. 
Such a consistent construction of the model ensures stable time gating in practice and is scalable to high-dimensional settings.

\subsection{Implementation details and model complexity.}

In our implementation, the PICNN depth and width are adapted to the problem scale. We use $L=2$ to $8$ hidden layers and hidden widths up to $1024$ in high-dimensional experiments, while smaller configurations are used for low-dimensional toy problems.
Each time-dependent bias network $S_l(t)$ is implemented as a three-layer MLP, and the scalar coefficient $\alpha(t)$ is generated by a similar MLP. All layers share the same hidden width for simplicity. The architecture depth and width depend on the problem scale (e.g., low-dimensional toy problems versus high-dimensional benchmarks).

 At the architecture level, the number of parameters scales as $\mathcal{O}(Lm^2 + Ldm)$, where $L$ is the number of PICNN layers, $m$ is the hidden width, and $d$ is the input dimension. This scaling includes both the PICNN layers and the time-dependent bias networks $S_l(t)$. In our experiments, $m$ is chosen according to the problem dimension and can be up to $1024$ for high-dimensional tasks, leading to model sizes on the order of $10^6$--$10^7$ parameters.

 During inference, a single evaluation of the velocity field requires one network forward pass and one automatic differentiation operation with respect to the input variable $x$, which remains comparable to standard neural-network evaluation complexity.

The overall training complexity additionally depends on stochastic sampling and loss approximation during optimization. Compared with OT-based FMs that require repeated minibatch OT matching or inner optimization procedures, the proposed framework avoids costly nested OT optimization during training. Following common practice, we therefore evaluate the practical computational scaling empirically. As reported in Table~\ref{tab:bm}, under the $D=256$ benchmark, COFM requires $28$K seconds, compared with $81$K seconds for OFM ($B=128$) and $254$K seconds for OFM ($B=1024$), while maintaining competitive or improved transport accuracy.

\subsection{Training: Minibatch pairing and subproblem-free consistency.}

At each iteration we sample $\{z_i\}_{i=1}^B\sim \mu_{0}$ and $\{x_i\}_{i=1}^B\sim \mu_{1}$, where $B$ denotes the size of mini-batch. By default, we form index-wise random pairings $(z_i,x_i)$. For applications, we optionally enable minibatch OT matching inspired from OT-CFM \cite{tong2024improving}, where a discrete OT problem with quadratic cost
 $C_{ij}=\frac{1}{2}\left\|z_{i}-x_{j}\right\|^{2}$ is solved to obtain batch-wise pairings between samples. In all experiments, we take $B=1024$ to balance the computational cost of network backpropagation and the discrete OT solver (implemented in the {\sc POT} package).

We then construct the displacement point $x_{t}^{\pi}=(1-t) z_{i}+t x_{i}$ and minimize a time-weighted flow-matching loss against the displacement velocity together with an HJ consistency term, both evaluated at $x_{t}^{\pi}$, the total loss is obtained by \eqref{eq:wholeloss}. All spatial derivatives of the OT potential and the HJ residual terms in (20) are computed using automatic differentiation.
For fairness, baseline methods (e.g., OFM) are trained under the \emph{same} minibatch size, ensuring comparable compute and memory budgets while highlighting the effect of our potential-based and temporally consistent training.

\subsection{Connection to optimal transport}
\label{subsec:ot-connection}
In this section, we present the relation between the loss function \eqref{eq:wholeloss}  and the optimal transport map in the $\Wtwo$ metric.
Under standard assumptions, the endpoint map $x\mapsto\nabla\Psi^\theta(0,x)$ coincides with the Brenier map. Thus, while our objective reduces to the familiar OFM alignment at $t=0$, the HJ identity extends this structure to all $t\in[0,1]$, promoting geodesic, structure-preserving evolution throughout. In implementation, we realize $x\mapsto\Psi^\theta(t,x)$ with a time-conditioned PICNN to enforce convexity uniformly in $t$ and to improve numerical conditioning near $t\to1$.

At points of differentiability, the $c^{s,t}$-subdifferential takes the explicit form
\[
\partial_{c^{s,t}}\psi(s,x)=\{\,x+(t-s)\nabla\psi(s,x)\,\},
\]
then optimal displacements are straight segments with velocity $\nabla\psi(s,x)$. In particular, samples evolve along straight lines $x_t=(1-t)x_0+tx_1$.
Additionally, with the rescaling \eqref{eq:scalepsi}
aligning $\Psi$ with \eqref{eq:HJequation} is exactly the consistency condition that matches the learned trajectories to displacement interpolation while preserving convex structure and the terminal anchor $\Psi^\theta(1,x)=\tfrac{1}{2}\|x\|^2$.

\begin{lem}[Parametric Fenchel-Young identities]\label{lem:param-FY}
Fix $t\in[0,1]$. Assume $\Psi^\theta(t,\cdot):\R^d\to\R\cup\{+\infty\}$ is proper, lower semi-continuous, and convex.
Let the convex conjugate with respect to the $x$-variable be
\[
\bar\Psi^{\theta}(t,y):=\sup_{z\in\R^d}\{\langle y,z\rangle-\Psi^\theta(t,z)\},\qquad y\in\R^d.
\]
Then for any $x,y\in\R^d$, if $\Psi^\theta(t,\cdot)$ is differentiable at $x$ and $\Psi^{\theta,*}(t,\cdot)$ is differentiable at $y$, then the subgradients reduce to gradients and the following statements are equivalent:
\begin{enumerate}
\item $y=\nabla_x \Psi^\theta(t,x)$;
\item $x\in \argmax_{z\in\R^d}\{\langle y,z\rangle-\Psi^\theta(t,z)\}$;
\item $\Psi^\theta(t,x)+\bar\Psi^{\theta}(t,y)=\langle x,y\rangle$;
\item $x=\nabla_y \bar\Psi^{\theta}(t,y)$;
\item $y\in\argmax_{z\in\R^d}\{\langle z,x\rangle-\bar\Psi^{\theta}(t,z)\}$.
\end{enumerate}
\end{lem}
\begin{proof}
  The Fenchel--Young inequality gives
\(
\Psi^\theta(t,x)+\bar\Psi^{\theta}(t,y)\ge \langle x,y\rangle
\)
for all $x,y$, with equality iff $y\in\partial_x\Psi^\theta(t,x)$,
which proves \((a)\Leftrightarrow(c)\).
Optimality in the concave maximization problems in (2) and (5)
is equivalent to the subgradient inclusions
\(y\in\partial_x\Psi^\theta(t,x)\) and \(x\in\partial_y\bar{\Psi^{\theta}}(t,y)\),
respectively, hence \((a)\Leftrightarrow(2)\) and \((b)\Leftrightarrow(5)\).
By conjugation, \((a)\Leftrightarrow(b)\) (i.e.\ $\partial\Psi^\theta(t,\cdot)=(\partial\bar{\Psi^{\theta}}(t,\cdot))^{-1}$).
Under the differentiability assumptions, subgradients are singletons,
yielding the gradient statements (1) and (4).
\end{proof}

The following theorem establishes a theoretical connection between the HJ-consistent FM objective and the Kantorovich dual formulation of OT, thereby providing theoretical justification for the proposed regularization strategy.
\begin{theorem}[HJ consistency and OT duality]\label{the:mainot}
Let $\Psi^\theta:[0,1]\times\R^d\to\R$ be proper, lower semi-continuous, and convex in $x$ for each $t$. Assume $\Psi^\theta$ is a solution of \eqref{eq:HJequation}. For fixed $(x_0,x_1)$, let $x_t=(1-t)x_0+t x_1$ denote the pointwise FM residual
\[
r(t;x_0,x_1)
:= \nabla\Psi^\theta(t,x_t) - \frac{x_t-x_0}{t},
\] if $\mathcal{L}_{\mathrm{HJ}}(\Psi^\theta)=0$,
then the following exact identity holds
\begin{equation}
    \int_{0}^{1}\left\|r\left(t ; x_{0}, x_{1}\right)\right\|^{2} \rmd t=2\left[\Psi^{\theta}\left(0, x_{0}\right)+\bar{\Psi}^{\theta}\left(0, x_{1}\right)-\left\langle x_{0}, x_{1}\right\rangle\right]
\end{equation}
As a consequence, for any $p_0, p_1 \in\mathcal{P}_2(\R^d)$ and any transport plan $\pi \in \Pi(p_0,p_1)$, the FM loss satisfies
\begin{equation}
    \mathcal{L}_{\mathrm{FM}}\left(\Psi^{\theta}\right)=2 \E_{(X_0,X_1)\sim\pi}\left[\Psi^{\theta}\left(0, X_{0}\right)+\bar\Psi^{\theta}\left(0, X_{1}\right)-\left\langle X_{0}, X_{1}\right\rangle\right].
\end{equation}
Under the assumptions above, minimizing $\mathcal{L}_{\mathrm{COFM}}$ in \eqref{eq:wholeloss} is equivalent to minimizing the dual functional OT $\mathcal{L}_{\mathrm{OT}}$ defined in \eqref{eq:w2dual}, over the same convex potentials, i.e.,
\begin{equation}
    \underset{ \Psi^{\theta}(0, \cdot)\text { convex}}{\arg \min } \mathcal{L}_{\mathrm{COFM}}\left(\Psi^{\theta}\right)=\underset{ \Psi^{\theta}(0, \cdot)\text{ convex}}{\arg \min } \mathcal{L}_{\mathrm{OT}}\left(\Psi^{\theta}\right) .
\end{equation}
\end{theorem}
Theorem~\ref{the:mainot} provides the theoretical justification for introducing the HJ-consistent residual into the FM framework. 
In particular, it explains why enforcing temporal consistency of the convex potential aligns the learned dynamics with Wasserstein geodesics and OT duality. 
This result also motivates the use of a single time-dependent PICNN potential to simultaneously characterize both the transport map and the underlying flow dynamics.

We further relate the endpoint Kantorovich duality gap to errors along the path.  {The next theorem quantifies how the endpoint OT dual gap can be controlled by the FM error and the HJ residual accumulated along the trajectory. }

\begin{theorem}[Dual-gap bound with FM and HJ residual]
\label{thm:dual-gap-robust}
Let $\Psi^\theta$ be $C^1$ in $t$, $C^2$ and convex in $x$, with boundary condition $\Psi^\theta(1,x)=\tfrac12\|x\|^2$.
Assume $\|\nabla^2\Psi^\theta(t,x)\|_{\mathrm{op}}\le L_H$ for all $(t,x)$ and that $\Psi^\theta(0,\cdot)$ is $m$-strongly convex.
Let $(X_0,X_1)\sim\pi$ and define the linear bridge $X_t=(1-t)X_0+tX_1$.
Recall the FM velocity mismatch $e_{\mathrm{FM}}$ in \eqref{def:efm} and the HJ residual $\mathcal{R}$ in \eqref{def:R}.
Fix any $\delta\in(0,1)$ and suppose that $\|\nabla_x\mathcal R(t,x)\|\le L_R$ on $[0,1]\times\R^d$.
Then
\begin{equation}\label{eq:dual-gap-robust}
\begin{aligned}
&\mathbb{E}\!\left[\Psi^\theta(0,X_0)+\bar\Psi^\theta(0,\cdot)(X_1)-X_0\!\cdot\!X_1\right]
\\&\leq
\frac{2L_H^2}{m}\mathcal{L}_{\mathrm{FM}}(\Psi^\theta)+\frac{2}{m} \int_{0}^{1-\delta} \mathbb{E}\left\|\nabla_x \mathcal{R}\left(t, X_{t}\right)\right\|^{2}\mathrm{d}t,
\end{aligned}
\end{equation}
\end{theorem}
Theorem~\ref{thm:dual-gap-robust} shows that the proposed objective admits a quantitative stability interpretation. 
Specifically, the OT duality gap can be controlled by two components: the flow matching approximation error and the violation of the HJ consistency condition. This decomposition clarifies the distinct roles of the two losses used in COFM: the FM term encourages accurate transport dynamics, while the HJ residual regularizes the temporal consistency of the learned convex potential.

The bound also highlights a potential instability near the terminal time $t\to1$, which motivates the following practical treatment.
\begin{remark}
    The integral $\int_{0}^{1-\delta}(1-t)^{-2}\mathrm{d}t=1/\delta-1$ makes explicit the sensitivity near $t\to 1$; in practice one either excludes a small neighborhood of $t=1$ (our assumption) or re-weights/samples $t$ with a density proportional to $(1-t)^2$ to keep the constant finite.
\end{remark}

Furthermore, if the optimal coupling $\pi^*$ is tractable, we have the following quantitative control.
\begin{theorem}\label{thm3}
Let $(X_0,Y_1)\sim\pi^*$ be the optimal coupling between $\mu_0$ and $\mu_1$,
so that $X_0\sim\mu_0$, $Y_1\sim\mu_1$, and $T^*=\nabla\Psi^*$ satisfies $Y_1=T^*(X_0)$.
Assume $\Psi^\theta(0,\cdot)$ is $m$-strongly convex in $x$.
Then, combining the dual-gap bound in \eqref{eq:dual-gap-robust} with strong convexity yields
\begin{equation}\label{eq:final-vel-bound}
\begin{aligned}
    \mathbb{E}_{X_0\sim\mu_0}&\!\Big[\|\nabla\Psi^\theta(0,X_0)-T^*(X_0)\|^2\Big]
\\\;\le\;
&\frac{4L_H^2}{m^2}\mathcal{L}_{\mathrm{FM}}(\Psi^\theta) \;+\;
\frac{4}{m^2}\int_0^1\mathbb{E}\|\nabla_x\mathcal R(t,X_t)\|^2\,\mathrm{d}t,
\end{aligned}
\end{equation}
where $\bar\Psi$ denotes the convex conjugate of $\Psi$, and
$L_H$ is the uniform Hessian bound from Theorem~\ref{thm:dual-gap-robust} (i.e., $\|\nabla^2\Psi^\theta(t,x)\|\le L_H$).
\end{theorem}

Theorem~\ref{thm3} provides a direct interpretation of the practical impact of minimizing the proposed COFM objective. 
It shows that reducing both the FM loss and the HJ residual improves the approximation quality of the learned transport map toward the optimal OT map. 
This result offers theoretical support for the empirical observation that the proposed method produces more stable and geometrically consistent transport trajectories.

\section{Experimental Results}\label{sec5}

We evaluate our approach on three complementary settings: an illustrative 2D example (\S\ref{sub:toy}), an OT modeling benchmark on dimensional scalability of the models (\S\ref{suc:bm}), and a practical applicative benchmark on an unpaired image-to-image translation task in the latent space of a pre-trained ALAE autoencoder (\S\ref{sub:image}). 

Across all tasks, we adopt the consistency loss $\mathcal{L}_{\mathrm{HJ}}$ with task-specific realizations. 
In the 2D example, we use the Hamilton--Jacobi residual $\mathcal{L}_{\mathrm{res}}$ to emphasize geometric consistency of the flow. 
For the $\Wtwo$ benchmark and the ALAE translation task, we instead use the pushforward loss $\mathcal{L}_{\mathrm{pf}}$, which provides improved numerical stability in high-dimensional settings. 
As discussed earlier, $\mathcal{L}_{\mathrm{pf}}$ and $\mathcal{L}_{\mathrm{res}}$ characterize the same notion of consistency error from different perspectives.

Unless otherwise specified, all experiments are conducted using comparable optimization settings without additional stabilization strategies such as EMA updates. 
All experiments are performed on the same GPU model (NVIDIA RTX A6000, 48 GB).

\smallskip
\noindent {\textbf{Compared Methods}: 
We compare COFM with both reference OT solvers and representative flow matching-based generative models.} 
 {Specifically, the reference/baseline methods include:
(1) the classical OT solver MMv1 \cite{stanczuk2021wasserstein}, and 
(2) a simple linear baseline that matches the first and second moments between distributions. 
The learned flow matching-based competitors include Optimal Flow Matching (OFM) \cite{OFM}, Conditional Flow Matching (OT-CFM) \cite{tong2023conditional}, Rectified Flow (RF) \cite{liu2023flow}, and its conditional variant c-RF \cite{liu2022rectified}. 
For different experimental settings (toy problems, high-dimensional OT benchmarks, and image-to-image translation), we use appropriate subsets of these methods depending on the task.}

\subsection{Toy problems}\label{sub:toy}
We train the model to transport a Gaussian source distribution $p_0 = \mathcal{N}(0, 0.5 I)$ to a prescribed target distribution. 
The target distribution is a mixture of $8$ Gaussian components with equal weights, whose means are uniformly arranged on a circle of radius $D=5$, and each component has isotropic covariance $\sigma^2 I$ with $\sigma^2=0.15$. 
Samples from both source and target distributions are independently generated, with $10{,}000$ samples used for training.
\begin{figure*}
    \centering
   \includegraphics[width=\linewidth,trim=0 2.8cm 0 3cm,clip]{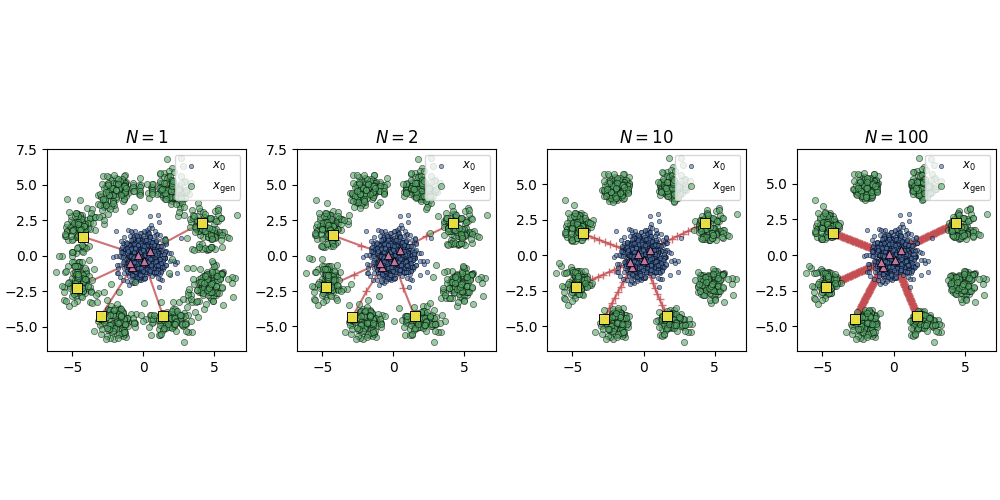}
   \caption{\textbf{Effect of ODE sampling steps.} Qualitative performance of a fixed trained model with \(N=1,2,10,100\) on the Gaussian to Eight-Gaussians setup in 2D. Five ($5$) realizations of the generation are included for illustration: from initial (triangle) to generation (square); $+$ denotes the intermediate steps of the ODE \eqref{eq:ode} generation $(N>1)$ with uniform time discretization. }
    \label{fig:8g}
\end{figure*}

As shown in Fig.~\ref{fig:8g}, to evaluate the sampling by \emph{one-step} and \emph{multi-step} regimes, we keep the
trained model $\Psi^\theta$ fixed and vary only the number of ODE integration steps
\(N=1,2,10,100\) at the sampling time. 
With \(N=1\), COFM already directs central mass to the correct modes, achieving good mode coverage and geometric alignment. Increasing the number of steps to $N=2$ and beyond further under ODE scheme \eqref{eq:ode} sharpens the clusters and reduces spurious outliers, yielding a closer match to the target distribution. For $N=10$ and $N=100$, the transport trajectories become nearly radial, indicating a vector field close to the displacement direction; multi-step integration mainly refines the already accurate one-step mapping. 

In summary, our approach establishes a \emph{consistency} model based on OT flow: it performs well in the \emph{one-step} regime and improves monotonically as $N$ increases, with no additional training cost (sampling-only change).

\subsection{High-dimensional OT benchmarks}\label{suc:bm}

We use the standard high-dimensional continuous OT benchmark from \cite{korotin2021bm}, where the source and target distributions are paired synthetic distributions in various dimensions. Since closed-form OT maps are generally unavailable in high-dimensional settings, the benchmark first employs a pre-trained time-independent ICNN solver to compute a numerical approximation of the Brenier transport map between the two distributions. This approximate OT solution is then treated as a reference transport map for evaluation.

Training samples are independently drawn from the source and target distributions, and no paired correspondence between samples is provided during training. Consequently, the benchmark evaluates how well learned transport models recover the transport geometry induced by the reference OT solver. Therefore, map-based metrics assess not only whether the transported distribution matches the target distribution, but also whether the learned transport trajectories remain consistent with the reference OT behavior.

Following \cite{korotin2021wasserstein}, we report the $\mathcal{L}^2$ unexplained variance percentage ($\mathcal{L}^2$--UVP) and a cosine similarity metric between transport directions. Since the ground-truth transport map is available in this benchmark, map-based metrics provide a direct and informative evaluation of transport accuracy.

Let $\|f\|_{L^2(p_0)}^2 := \mathbb{E}_{x\sim p_0}\!\left[\|f(x)\|^2\right]$, for a learned map $T$,
\[
\mathcal{L}^2\text{–UVP}(T)
:= 100 \cdot \frac{\|T - T^*\|_{L^2(p_0)}^{2}}{\mathrm{Var}(p_1)} \ \% ,
\]
where $\mathrm{Var}(p_1)$ denotes the total variance of $p_1$. Smaller values are better, $\mathcal{L}^2\text{–UVP}(T)\to 0\%$ indicates $T\to T^*$.

 {We compare COFM with several representative and SOTA methods for learning transport maps, including} OFM\cite{OFM}, OT-CFM\cite{tong2023conditional}, RF\cite{liu2023flow}, and c-RF\cite{liu2022rectified}.
We also include the classical dual OT solver MMv1 \cite{stanczuk2021wasserstein} as a reference, as well as a simple linear map that matches the first and second moments between $p_0$ and $p_1$.

Table~\ref{tab:bm} enumerates the compared solvers, including several flow-matching baselines and two non-generative OT solvers. In particular, MMv1 is a dual OT solver and linear is a baseline solver; neither produces samples, although both are very fast. We nevertheless report linear as a \emph{reference baseline}: if a learned solver achieves a $L^2$-UVP lower than the linear baseline, it indicates the ability to approximate the OT mapping rather than merely fitting moments. To align the settings between methods, we follow the OFM experimental protocol and the public implementation (\href{https://github.com/Jhomanik/Optimal-Flow-Matching}{https://github.com/Jhomanik/Optimal-Flow-Matching}
) and use the same hardware. For all reported methods, we fix the learning rate to $10^{-3}$ in the training stage to ensure comparability. COFM is trained with a batch size of $B=1024$ for $200$K iterations. For \emph{OFM}, we report results at both $B=128$ and $B=1024$, following the original $30$K training steps. On high-dimensional benchmarks, increasing $ B$ to $ 1024$ does not produce a noticeable improvement in $L^2$-UVP over $B=128$, while incurring substantially higher training costs. More precisely,  we record the training time of two OT solvers under $D=256$. OFM takes $81,067$s with $B=128$ and $254,472$s with $B=1024$  for $30\mathrm{k}$ training iterations; COFM  with $B=1024$ takes $28{,}788$s for $200\mathrm{k}$ training iterations. Despite using more steps, our per-step computation is lightweight, keeping the total runtime low. We further tested with increasing training steps, and the performance of each method did not improve significantly. 

\begin{figure}
    \centering
    \includegraphics[width=1\linewidth]{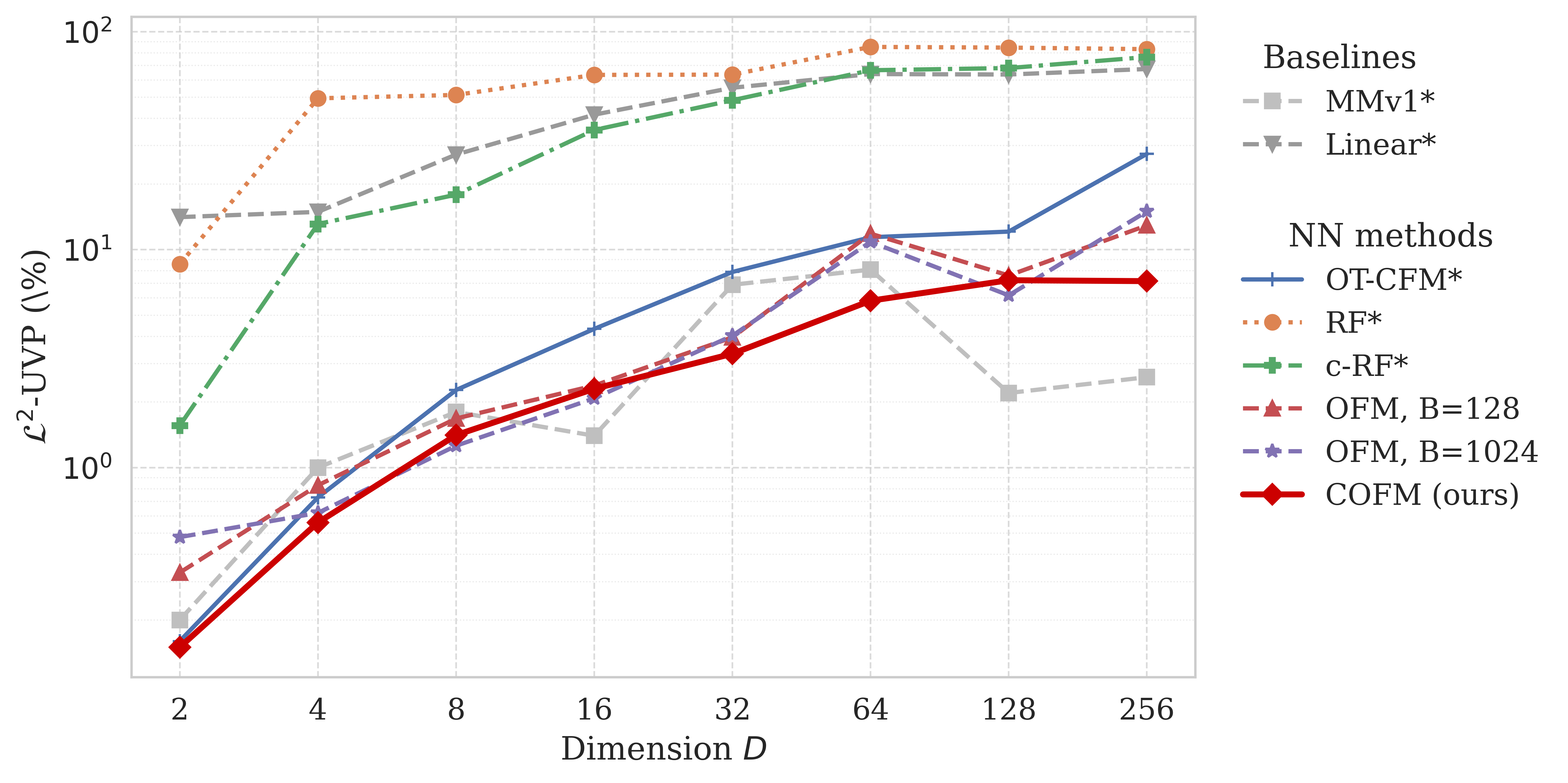}
    \caption{Comparison of solvers across different dimensions $D$. *Metrics are taken from \cite{OFM}.}
    \label{figsolver-bm}
\end{figure}

As shown in Table~\ref{tab:bm} and Figure~\ref{figsolver-bm}, COFM stands out in the high-dimensional regime; it attains the best among all flow-matching–based solvers. Equally notable is its stability, as the dimension increases from $2$ to $256$, our metric rises smoothly from $0.15$ to $7.17$ with near-linear growth and no spikes or oscillations. The trend lines in Figure~\ref{figsolver-bm} corroborate this behavior-ours (red curve) stays consistently near the bottom and increases gently with $D$, whereas RF/c\mbox{-}RF deteriorates sharply and OFM exhibits pronounced high-dimensional fluctuations. In addition, our training avoids the inner optimization subproblems \eqref{ofm-ot}, resulting in a shorter wall-clock time (as mentioned, $28$K seconds v.s. $254$K seconds) and better accuracy in high dimensions than OFM under identical hardware and batch sizes.

 {To further assess the robustness of the proposed method, we additionally report the mean and standard deviation of COFM over multiple random seeds in Table \ref{tab:bm}. The relatively small variance across different dimensions demonstrates that the proposed framework exhibits stable optimization behavior and consistent transport performance. In particular, the variance remains well controlled even in high-dimensional settings, indicating the robustness of the convex-potential parameterization and the proposed consistency regularization.}

\begin{table*}[ht]
\centering
\caption{$\mathcal{L}^2$-UVP$\downarrow$ values of solvers fitted on high-dimensional benchmarks in various dimensions.
The best metric over the Flow Matching solvers is highlighted. *Metrics are taken from \cite{OFM}. Computational time of the last three OT solvers under $D=256$: OFM ($B=128$), $81$K seconds; OFM ($B=1024$), $254$K seconds; COFM $28$K seconds. For COFM, we additionally report the mean $\pm$ standard deviation over repeated runs with different random seeds.}
\renewcommand{\arraystretch}{1}
\setlength{\tabcolsep}{3pt}
\begin{tabular}{llcccccccccc}
\toprule
\textbf{Solver} & \textbf{Type} & $D=2$ & $D=4$ & $D=8$ & $D=16$ & $D=32$ & $D=64$ & $D=128$ & $D=256$ \\
\midrule
MMv1\textsuperscript{*}  & Dual OT  & 0.2 & 1.0 & 1.8 & 1.4 & 6.9 & 8.1 & 2.2 & 2.6  \\
Linear\textsuperscript{*}  & Baseline & 14.1 & 14.9 & 27.3 & 41.6 & 55.3 & 63.9 & 63.6 & 67.4 \\
\midrule
OT-CFM\textsuperscript{*} & \multirow{6}{*}{FM} & 0.16 & 0.73 & 2.27 & 4.33 & 7.9 & 11.4 & 12.1 & 27.5 \\
RF\textsuperscript{*} && 8.58 & 49.46 & 51.25 & 63.33 & 63.52 & 85.13 & 84.49 & 83.13 \\
c-RF\textsuperscript{*} & & 1.56 & 13.11 & 17.87 & 35.39 & 48.46 & 66.52 & 68.08 & 76.48 \\
OFM  $B=128$  & & 0.33 & 0.83 & 1.68 & 2.38 & 3.96 & 11.84 & 7.61 & 12.92 \\
OFM  $B=1024$  & & 0.48 & 0.62 & \textbf{1.26} & \textbf{2.08} & 4.03 & 10.86 & \textbf{6.17} & 14.98 \\

{{\shortstack{\textbf{COFM (ours)}\\ {\textit{(mean $\pm$ std)}}}}} & &

\multicolumn{1}{c}{{\shortstack{\textbf{0.15}\\ \textit{$\pm$0.0004}}}} &
\multicolumn{1}{c}{{\shortstack{\textbf{0.56}\\ \textit{$\pm$0.0080}}}} &
\multicolumn{1}{c}{{\shortstack{1.41\\ \textit{$\pm$0.0001}}}} &
\multicolumn{1}{c}{{\shortstack{2.30\\ \textit{$\pm$0.0640}}}} &
\multicolumn{1}{c}{{\shortstack{\textbf{3.33}\\ \textit{$\pm$0.1091}}}} &
\multicolumn{1}{c}{{\shortstack{\textbf{5.83}\\ \textit{$\pm$0.0192}}}} &
\multicolumn{1}{c}{{\shortstack{7.24\\ \textit{$\pm$0.1205}}}} &
\multicolumn{1}{c}{{\shortstack{\textbf{7.17}\\ \textit{$\pm$0.5781}}}} \\

\bottomrule
\end{tabular}
\label{tab:bm}
\end{table*}

\smallskip
\noindent\textbf{ {Ablation study}} 
 {To further evaluate the effectiveness of the proposed framework, we conduct an ablation study on high-dimensional OT benchmark under the same training budget (50k iterations). The results are reported in Tabel~\ref{tab:ablation}.}

\begin{table*}[ht]
\centering
\caption{Ablation study on the high-dimensional OT benchmark (batch size=1024).
The variant ``FM only'' removes both $\mathcal{L}_{\mathrm{res}}$ and $\mathcal{L}_{\mathrm{pf}}$ and is trained only with the flow-matching loss. 
Lower $\mathcal{L}^2$-UVP indicates better performance.}
\label{tab:ablation}
\renewcommand{\arraystretch}{1}
\setlength{\tabcolsep}{4pt}
\small
\begin{tabular}{lccc cccccccc}
\toprule
\textbf{Variant} 
& \textbf{ Potential} 
& \textbf{FM} 
& \textbf{ Consistency} 
& \multicolumn{8}{c}{$\boldsymbol{\mathcal{L}^2}$\textbf{-UVP} $\downarrow$} \\
\cmidrule(lr){5-12}
& & & 
& $D=2$ & $4$ & $8$ & $16$ 
& $32$ & $64$ & $128$ & $256$ \\
\midrule
FM only 
& PICNN 
& \checkmark 
& -- & 0.43 
& 1.26 & 2.89 & 5.54 & 10.04 & 12.30 &  14.25 & 16.19 \\

w/o convexity 
& MLP 
& \checkmark 
& $\mathcal{L}_{\mathrm{pf}}$ 
& 17.65 & 21.93 & 9.38 & 10.44 & 6.21 & 14.73 & 9.02 & 13.72 \\

COFM-res 
& PICNN 
& \checkmark 
& $\mathcal{L}_{\mathrm{res}}$ 
& 0.32 & 1.11 & 2.64 & 3.78 & 6.67 & 11.94 & 9.24 & 17.78 \\

COFM-pf 
& PICNN 
& \checkmark 
& $\mathcal{L}_{\mathrm{pf}}$ 
& 0.15 & 0.56 & 1.41 & 2.30 & 3.33 & 5.83 & 7.24 & 7.17 \\
\bottomrule
\end{tabular}
\end{table*}

We first show the role of the convex-potential parameterization. Replacing the PICNN with a standard MLP (“w/o convexity”) leads to unstable optimization behavior, particularly in low-dimensional settings. For example, the $\mathcal{L}^{2}$-UVP increases from $0.15$ to $17.65$ in the two-dimensional experiment. Although the MLP-based model rapidly decreases training loss during the early stage, its transport quality deteriorates after prolonged training and exhibits large fluctuations. This suggests that unconstrained parameterizations may overfit the flow-matching objective without preserving stable transport geometry. Overall, these observations indicate that the combination of convex potential parameterization with the proposed HJ regularization improves the transport stability in our framework. 

Next, we evaluate the effect of consistency regularization. The "FM only" variant removes both consistency terms and is trained solely with the FM loss. Although it achieves reasonable performance in low-dimensional settings (e.g. $0.43$ at $D=2$), the transport error progressively increases as the dimensionality increases, reaching $16.19$ at $D=256$. This suggests that additional HJ-based regularization is necessary to align the learned dynamics with the underlying OT structure in high-dimensional settings.

Finally, we compare two implementations of the proposed consistency mechanism:  $\mathcal{L}_{\mathrm{res}}$ and $\mathcal{L}_{\mathrm{pf}}$. Both variants consistently outperform the FM-only baseline across most dimensions. For instance, at $D=16$, the $\mathcal{L}^{2}$-UVP decreases from $5.54$ (FM only) to $3.78$ with $\mathcal{L}_{\mathrm{res}}$, and further to $2.30$ with $\mathcal{L}_{\mathrm{pf}}$. Similarly, at $D=128$, the error is reduced from $14.25$ to $9.24$ and $7.24$, respectively. Although the two formulations are equivalent at the continuous formulation level under the assumption of Theorem \ref{the:mainot}, their numerical behaviors differ in practice. One possible explanation is that, in hign-dimensional settings, the displacement interpolation paths occupy only a small portion of the effective transport domain. As a result, the residual-based consistency term mainly constrains local trajectory behavior near sampled interpolation paths, whereas the pushforward formulation imposes a more global constraint. Such that $\mathcal{L}_{\mathrm{pf}}$ achieves improved empirical performance on high-dimensional benchmarks. 

\subsection{Image to image transfer}\label{sub:image}
Unpaired image-to-image translation is another task that learns a mapping between two distributions.  {We compare our method with representative flow-based baselines (OT-CFM and OFM) under the same one-step sampling setting.} We perform the translation in the 512-dimensional latent space of a pre-trained ALAE model trained on the FFHQ dataset at 1024×1024 resolution \cite{pidhorskyi2020adversarial}. Specifically, we split the 60k FFHQ training faces into male and female subsets and treat their ALAE latent codes as the source and target distributions $p_0$ and $p_1$. During inference, a male face from the FFHQ test set is encoded into latent space, transported by the learned model, and decoded back to image space.
  \begin{figure*}[!t]
  \centering
  \setlength{\tabcolsep}{0pt}
  \renewcommand{\arraystretch}{0}
  \begin{minipage}[t]{12mm}
           \begin{tabular}{@{}m{10mm}@{}}
      \vtag{}{$x\sim p_0$} \\[8mm]
      \vtag{Ours}{$x\sim p_1$}    \\[8mm]
      \vtag{OT-CFM}{$x\sim p_1$}     \\[10mm] 
      \vtag{OFM}{$x\sim p_1$}     \\[10mm]  
      \vtag{}{$x\sim p_0$} \\[8mm]
      \vtag{Ours}{$x\sim p_1$}    \\[8mm]
      \vtag{OT-CFM}{$x\sim p_1$}     \\[10mm] 
      \vtag{OFM}{$x\sim p_1$}
    \end{tabular}
\end{minipage}
\begin{minipage}[t]{0.9\linewidth}
  \begin{tabular}{@{}c@{}}
    \includegraphics[width=\linewidth,clip,trim=0 6.5 0 2]{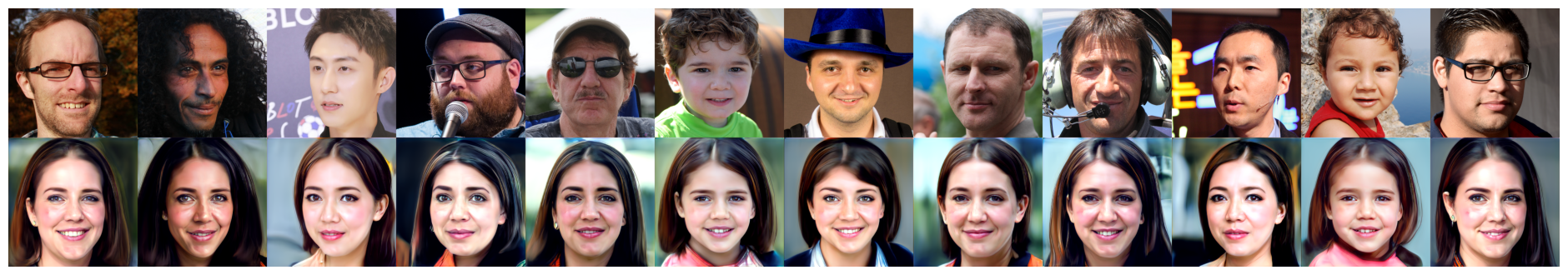}\\[-0.5mm]
        \includegraphics[width=0.99\linewidth,clip,trim=0 0 0 6.5]{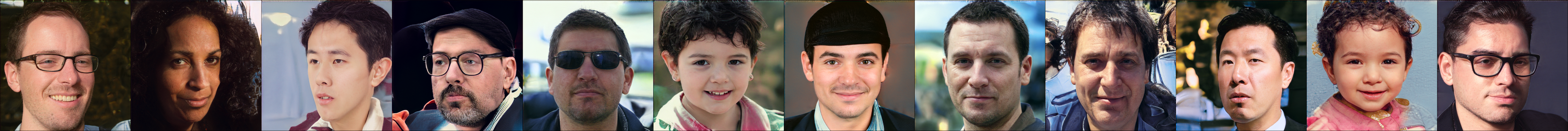}  \\[-0.5mm]
    \includegraphics[width=\linewidth,clip,trim=0 0 0 6.5]{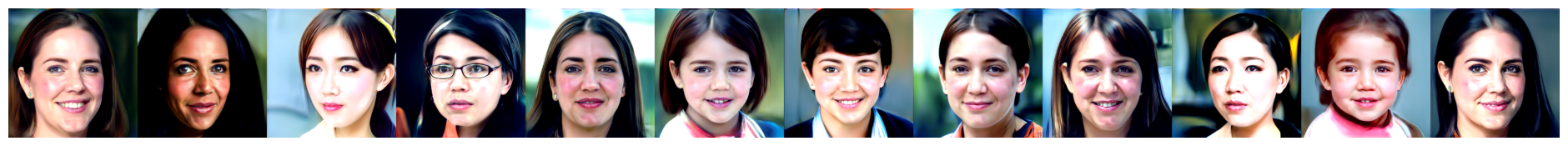}  \\
    \includegraphics[width=\linewidth,clip,trim=0 6.5 0 2]{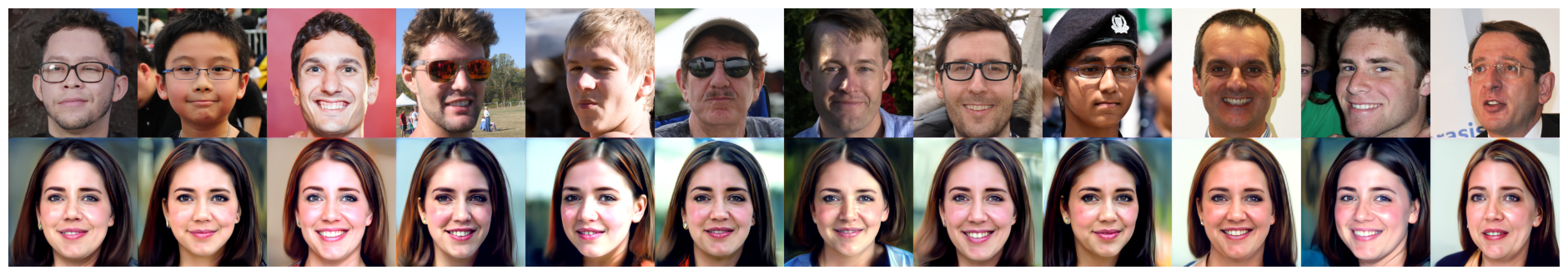} \\[-0pt]
        \includegraphics[width=0.99\linewidth,clip,trim=0 0 0 6.5]{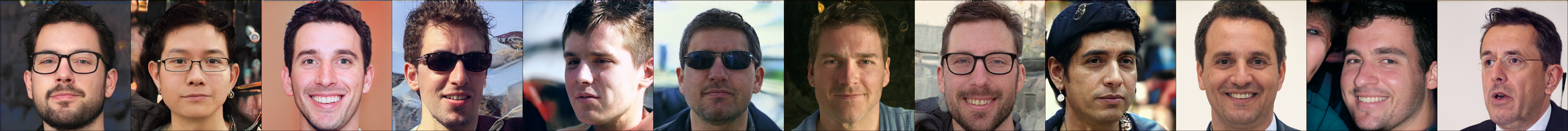}  \\[-0.5mm]
    \includegraphics[width=\linewidth,clip,trim=0 0 0 6.5]{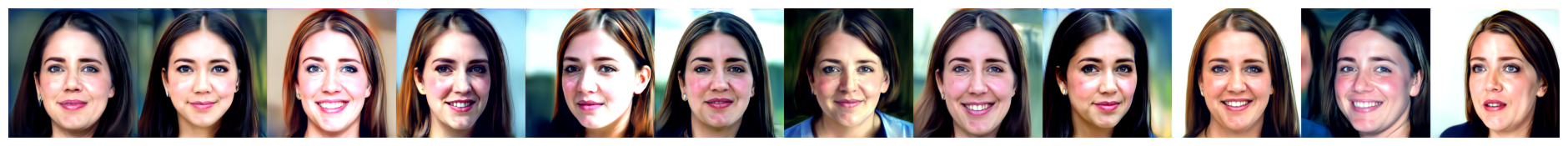} 
  \end{tabular}
\end{minipage}
  \caption{Unpaired image to image translation (male→female) using our COFM solver and OFM solver in the FFHQ $1024\times1024$ ALAE latent space. ($N=1$ of all the above results).}
  \label{fig:man2woman}
\end{figure*}
Under a one-step sampling regime (NFE=1) shared by all methods, our solver produces the most convincing male $\to$ female translations in Fig.~\ref{fig:man2woman}. Compared to OFM and OT--CFM, our results preserve identity geometry (pose, eyewear, jawline) while naturally injecting gender attributes. Textures are sharper with cleaner hair strands and more consistent skin shading; colors remain balanced rather than over-smoothed. Our results are more consistent and exhibit fewer artifacts. These observations indicate that our one--step map is closer to the desired OT solution, achieving higher perceptual fidelity under an extremely low compute budget.

\section{Conclusion}\label{sec6}
We proposed a machine learning framework for solving the optimal transport plan between two continuous distributions from their i.i.d samples, referred to as COFM. The proposed framework employs a time-dependent convex potential parameterized by a PICNN, which defines a conservative velocity field $v_t$. A HJ residual regularizes the dynamics toward constant-speed $\Wtwo$ geodesics. This formulation reduces the reliance on costly minibatch OT optimization during training, yielding a single training objective that supports both one-step and multi-step ODE sampling from the same potential. We prove a dual-gap bound that separates the contributions of the flow-matching error and the HJ residual, with an endpoint-robust truncated form and an endpoint-consistent form when the network enforces $\mathcal{O}(1-t)$ behavior near $t=1$. Empirically, the method trains stably, scales well to higher dimensions, and improves as the number of integration steps increases.

 {Despite these advantages, the current framework has several limitations. Our analysis currently assumes a quadratic cost and relies on bounded endpoint behavior, which may restrict applicability in more general optimal transport settings. In addition, the expressivity of the model depends on the capacity of the PICNN and the conservative field parameterization. Moreover, the convexity constraint may limit its ability to represent highly non-convex or multi-modal transport maps. Moreover, the inclusion of the HJ residual introduces additional computational overhead due to the evaluation of temporal derivatives. Finally, the current formulation focuses on deterministic transport and does not explicitly capture stochasticity in the dynamics.}

 Future directions include extending the proposed framework to entropic regularization and stochastic transport formulations, such as Schr\"odinger bridge models \cite{de2021diffusion,chen2016relation}. Another promising direction is the investigation of adaptive time parameterization and endpoint stabilization mechanisms for improving transport efficiency and trajectory consistency. We also plan to explore larger-scale generative applications, including high-resolution image synthesis \cite{esser2024scaling,rombach2022high} and conditional generative modeling\cite{lipmanflow}. These directions may further improve the flexibility, scalability, and applicability of geometry-aware flow matching frameworks.

\appendix
\noindent


\section{{Proofs of Theorems}}\label{app:proof}
Here we present detailed proofs.

\noindent{\textbf{Proof of Theorem \eqref{the:mainot} HJ consistency and OT duality.}}
\begin{proof}
For each $t\in(0,1)$, there exists a minimizer $z_0(t)$ such that
\[
X_{t}=(1-t) z_{0}(t)+t \nabla \Psi^{\theta}\left(0, z_{0}(t)\right).
\]
Consequently, 
\begin{equation}\label{eq:OTcon}
    \int_{0}^{1}\left\|r\left(t ; x_{0}, x_{1}\right)\right\|^{2}\rmd t=\int_{0}^{1} \frac{\left\|z_{0}(t)-x_{0}\right\|^{2}}{t^{2}} \rmd t.
\end{equation}
From the affine relation above, we also have $t\left(\nabla \Psi^{\theta}\left(0, z_{0}(t)\right)-x_{1}\right)=(1-t)\left(x_{0}-z_{0}(t)\right),$ so that
\[
\left\|\nabla \Psi^{\theta}\left(0, z_{0}(t)\right)-x_{1}\right\|^{2}=\frac{(1-t)^{2}}{t^{2}}\left\|z_{0}(t)-x_{0}\right\|^{2}.
\]
With the change of variables $s=t/(1-t)$, the right side of \eqref{eq:OTcon} becomes
\begin{equation}\label{eq:2}
    \int_{0}^{\infty}\left\|\nabla \Psi^{\theta}\left(0, z_{0}(s)\right)-x_{1}\right\|^{2} \rmd s.
\end{equation}
Using the identity $s(\nabla\Psi^\theta(0,z_0(s))-x_1)=x_0-z_0(s)$ yields
\begin{equation}\label{eq:3}
\begin{aligned}
    &\int_{0}^{\infty} \left\lVert \nabla\Psi^{\theta}\bigl(0,z_{0}(s)\bigr) - x_{1} \right\rVert^{2} \rmd s
\\= &2\Bigl[ \Psi^{\theta}(0,x_{0}) + \bar{\Psi}^{\theta}(0,x_{1}) - \langle x_{0},x_{1} \rangle \Bigr].
\end{aligned}
\end{equation}
Combining \eqref{eq:2} with \eqref{eq:3} proves the pathwise identity
\[
\int_{0}^{1} \left\lVert r(t; x_{0}, x_{1}) \right\rVert^{2} \rmd t
= 2\Bigl[ \Psi^{\theta}(0,x_{0}) + \bar{\Psi}^{\theta}(0,x_{1}) - \langle x_{0}, x_{1} \rangle \Bigr].
\]
Taking expectation over any coupling $\pi\in\Pi(p_0,p_1)$ gives
\begin{equation*}
\begin{aligned}
     \mathcal{L}_{\mathrm{FM}}(\Psi^{\theta})
&= \int_{0}^{1} \mathbb{E}_{(X_0,X_1)\sim\pi}\!\left[ \| r(t; X_{0}, X_{1}) \|^{2} \right] \rmd t
\\&= 2 \mathbb{E}_{(X_0,X_1)\sim\pi}\!\left[ \Psi^{\theta}(0, X_{0}) + \bar{\Psi}^{\theta}(0, X_{1}) - \langle X_{0}, X_{1} \rangle \right], 
\end{aligned}
\end{equation*}
so minimizing $ \mathcal{L}_{\mathrm{FM}}$ under $\mathcal{L}_{\mathrm{HJ}}(\Psi^\theta)=0$ is equivalent to minimizing the OT dual functional.
\end{proof} 

\noindent{\textbf{Proof of Theorem \eqref{thm:dual-gap-robust}: Dual-gap bound with FM and HJ residual.}}
\begin{proof}
Set $g(t):=\nabla\Psi^\theta(t,X_t)-X_1$ By the  definition of $\mathcal{R}$, we have
$$\nabla_x\mathcal{R}(t,x)=\partial_t\nabla\Psi^\theta(t,x)+\frac{1}{1-t}\nabla^2\Psi^\theta(t,x)(\nabla\Psi^\theta(t,x)-x).$$
Combining the above and applying the chain rule of $g$, for $t\in[0,1-\delta], $
we obtain 
\[
\left\|g^{\prime}(t)\right\| \leq \left\|\nabla_x \mathcal{R}\left(t, X_{t}\right)\right\|+L_{H}\left\|e_{\mathrm{FM}}\left(t, X_{t}\right)\right\|.
\]
Since $g(1)=0$ and $g(1-\delta)=-e_{\mathrm{FM}}(1-\delta)$, for any $\delta\in(0,1)$,
\begin{align}\label{eq:g0}
    g(0)=-\int_{0}^{1-\delta} g^{\prime}(t)\,\mathrm{d}t+g(1-\delta)
\end{align}
Using $(a+b)^2\le 2a^2+2b^2$, we obtain
\[
\E\|g(0)\|^{2} \le 2\,\mathbb{E}\Big\|\int_{0}^{1-\delta} g^{\prime}(t) \mathrm{d}t\Big\|^{2}
+2\, \E\|g(1-\delta)\|^{2},
\]
then
\begin{align*}
 \E\|g(0)\|^{2} \leq 4 \int_{0}^{1-\delta} \mathbb{E}\left\|\nabla_{x} \mathcal{R}\left(t, X_{t}\right)\right\|^{2} d t+6 L_{H}^{2} \mathcal{L}_{\mathrm{FM}}(\Psi^\theta)
\end{align*}
If $\|\nabla\mathcal{R}(t,x)\| \le L_{R}$ uniformly on $[0,1-\delta]$, then
\[
\int_{0}^{1-\delta} \mathbb{E}\|\nabla\mathcal{R}(t,X_t)\|^{2} \rmd t
\le L_R^2,
\]
by Lemma~\ref{lem:param-FY} and $m$-strong convexity of $\Psi^\theta(0,\cdot)$,
the endpoint Fenchel--Young gap satisfies
\[
\Psi^\theta(0,x)+\bar\Psi^\theta(0,\cdot)(y)-x\!\cdot\!y
\le \frac{1}{2m}\,\|y-\nabla\Psi^\theta(0,x)\|^2.
\]
With $x=X_0$, $y=X_1$ we obtain
\[
\mathbb{E}\big[\Psi^\theta(0,X_0)+\bar\Psi^\theta(0,\cdot)(X_1)-X_0\!\cdot\!X_1\big]
\le \frac{1}{2m}\,\mathbb{E}\|g(0)\|^{2},
\]
which combined with the preceding bounds yields \eqref{eq:dual-gap-robust}.
\end{proof}

\noindent{\textbf{Proof of Theorem\eqref{thm3}:}}
\begin{proof}
Let $\Psi=\Psi^\theta(0,\cdot)$. For $m$-strongly convex $\Psi$, by Lemma~\ref{lem:param-FY} for all $u,v\in\mathbb{R}^d$:
\begin{equation}\label{eq:FY-strong}
\Psi(u)+\bar\Psi(v)-\langle u,v\rangle \;\ge\; \frac{m}{2}\,\|\nabla\Psi(u)-v\|^2.
\end{equation}
Apply \eqref{eq:FY-strong} with $u=X_0$ and $v=T^*(X_0)$, take expectation over $X_0\sim\mu_0$,
and use $T^*_{\#}\mu_0=\mu_1$ to rewrite
$\mathbb{E}_{X_0\sim\mu_0}\bar\Psi\!\big(T^*(X_0)\big)=\mathbb{E}_{Y_1\sim\mu_1}\bar\Psi(0,Y_1)$.
We obtain the gap-to-velocity inequality
\begin{equation}\label{eq:gap-to-vel}
\begin{aligned}
&\mathbb{E}_{X_0\sim\mu_0}\!\Big[\|\nabla\Psi^\theta(0,X_0)-T^*(X_0)\|^2\Big]
\\&\;\le\; \frac{2}{m}\Big(
\mathbb{E}_{X_0\sim\mu_0}\Psi^\theta(0,X_0)
+\mathbb{E}_{Y_1\sim\mu_1}\bar\Psi^\theta(0,Y_1)
\\&-\mathbb{E}_{(X_0,Y_1)\sim\pi^*}\langle X_0,Y_1\rangle
\Big).
\end{aligned}
\end{equation}
Finally, substitute the dual-gap estimate \eqref{eq:dual-gap-robust} (from Theorem~\ref{thm:dual-gap-robust}) into the
right-hand side of \eqref{eq:gap-to-vel} to obtain \eqref{eq:final-vel-bound}.
\end{proof}

\bibliographystyle{alpha}
\bibliography{sample}

@book{villani2009optimal,
  title={Optimal transport: old and new},
  author={C{\'e}dric Villani},
  volume={338},
  year={2009},
  publisher={Springer}
}

@inproceedings{kingma2018glow,
  title={Glow: Generative flow with invertible 1x1 convolutions},
  author={Kingma, Durk P and Dhariwal, Prafulla},
  booktitle={Advances in Neural Information Processing Systems},
  volume={31},
  year={2018}
}

@article{froese2012numerical,
  title={A numerical method for the elliptic Monge--Amp\`{e}re equation with transport boundary conditions},
  author={Froese, Brittany D},
  journal={SIAM Journal on Scientific Computing},
  volume={34},
  number={3},
  pages={A1432--A1459},
  year={2012},
  URL = {https://doi.org/10.1137/110822372},
  publisher={SIAM}
}

@inproceedings{
korotin2021wasserstein,
title={Wasserstein-2 Generative Networks},
author={Alexander Korotin and Vage Egiazarian and Arip Asadulaev and Alexander Safin and Evgeny Burnaev},
booktitle={International Conference on Learning Representations (ICLR)},
year={2021},
url={https://openreview.net/forum?id=bEoxzW_EXsa}
}

@article{stanczuk2021wasserstein,
  title={Wasserstein GANs work because they fail (to approximate the Wasserstein distance)},
  author={Stanczuk, Jan and Etmann, Christian and Kreusser, Lisa Maria and Sch{\"o}nlieb, Carola-Bibiane},
  journal={arXiv preprint arXiv:2103.01678},
url={https://arxiv.org/abs/2103.01678},
  year={2021}
}

@article{froese2011convergent,
  title={Convergent finite difference solvers for viscosity solutions of the elliptic Monge--Amp{\`e}re equation in dimensions two and higher},
  author={Froese, Brittany D and Oberman, Adam M},
  journal={SIAM Journal on Numerical Analysis},
  volume={49},
  number={4},
  pages={1692--1714},
  year={2011},
  URL = {https://doi.org/10.1137/100803092},
  publisher={SIAM}
}

@inproceedings{makkuva2020optimal,
  title={Optimal transport mapping via input convex neural networks},
  author={Makkuva, Ashok and Taghvaei, Amirhossein and Oh, Sewoong and Lee, Jason},
  booktitle={International Conference on Machine Learning},
  pages={6672--6681},
  year={2020},
  url = 	 {https://proceedings.mlr.press/v119/makkuva20a.html},
  organization={PMLR}
}

@inproceedings{neuralODE,
  author = {Chen, Ricky T. Q. and Rubanova, Yulia and Bettencourt, Jesse and Duvenaud, David K},
 booktitle = {Advances in Neural Information Processing Systems},
 editor = {S. Bengio and H. Wallach and H. Larochelle and K. Grauman and N. Cesa-Bianchi and R. Garnett},
 pages = {},
 publisher = {Curran Associates, Inc.},
 title = {Neural Ordinary Differential Equations},
 url = {https://proceedings.neurips.cc/paper_files/paper/2018/file/69386f6bb1dfed68692a24c8686939b9-Paper.pdf},
 volume = {31},
 year = {2018}
}

@article{song2020score,
      title={Score-Based Generative Modeling through Stochastic Differential Equations}, 
      author={Yang Song and Jascha Sohl-Dickstein and Diederik P. Kingma and Abhishek Kumar and Stefano Ermon and Ben Poole},
      year={2021},
      eprint={2011.13456},
      archivePrefix={arXiv},
      primaryClass={cs.LG},
      journal={arXiv preprint},
      url={https://arxiv.org/abs/2011.13456}
}

@article{
tong2024improving,
title={Improving and generalizing flow-based generative models with minibatch optimal transport},
author={Alexander Tong and Kilian FATRAS and Nikolay Malkin and Guillaume Huguet and Yanlei Zhang and Jarrid Rector-Brooks and Guy Wolf and Yoshua Bengio},
journal={Transactions on Machine Learning Research},
issn={2835-8856},
year={2024},
url={https://openreview.net/forum?id=CD9Snc73AW},
note={Expert Certification}
}

@inproceedings{onken2021ot,
  title={Ot-flow: Fast and accurate continuous normalizing flows via optimal transport},
  author={Onken, Derek and Fung, Samy Wu and Li, Xingjian and Ruthotto, Lars},
  booktitle={Proceedings of the AAAI Conference on Artificial Intelligence},
  volume={35},
  number={10},
  pages={9223--9232},
  year={2021},
url={https://ojs.aaai.org/index.php/AAAI/article/view/17113}
}

@inproceedings{lipmanflow,
title={Flow Matching for Generative Modeling},
author={Yaron Lipman and Ricky T. Q. Chen and Heli Ben-Hamu and Maximilian Nickel and Matthew Le},
booktitle={International Conference on Learning Representations (ICLR)},
year={2023},
url={https://openreview.net/forum?id=PqvMRDCJT9t}
}

@inproceedings{gushchin2024light,
title={Light and Optimal Schr\"odinger Bridge Matching},
author={Nikita Gushchin and Sergei Kholkin and Evgeny Burnaev and Alexander Korotin},
booktitle={Forty-first International Conference on Machine Learning},
year={2024},
url={https://openreview.net/forum?id=EWJn6hfZ4J}
}

@inproceedings{gat2024discrete,
 author = {Gat, Itai and Remez, Tal and Shaul, Neta and Kreuk, Felix and Chen, Ricky T. Q. and Synnaeve, Gabriel and Adi, Yossi and Lipman, Yaron},
 booktitle = {Advances in Neural Information Processing Systems},
 editor = {A. Globerson and L. Mackey and D. Belgrave and A. Fan and U. Paquet and J. Tomczak and C. Zhang},
 pages = {133345--133385},
 publisher = {Curran Associates, Inc.},
 title = {Discrete Flow Matching},
 url = {https://proceedings.neurips.cc/paper_files/paper/2024/file/f0d629a734b56a642701bba7bc8bb3ed-Paper-Conference.pdf},
 volume = {37},
 year = {2024}
}

@article{yang2024consistency,
  title={Consistency Flow Matching: Defining Straight Flows with Velocity Consistency},
  author={Ling Yang and Zixiang Zhang and Zhilong Zhang and Xingchao Liu and Minkai Xu and Wentao Zhang and Chenlin Meng and Stefano Ermon and Bin Cui},
  journal={arXiv preprint arXiv:2407.02398},
  year={2024},
  url={https://arxiv.org/abs/2407.02398}
}

@inproceedings{OFM,
 author = {Kornilov, Nikita and Mokrov, Petr and Gasnikov, Alexander and Korotin, Alexander},
 booktitle = {Advances in Neural Information Processing Systems},
 editor = {A. Globerson and L. Mackey and D. Belgrave and A. Fan and U. Paquet and J. Tomczak and C. Zhang},
 pages = {104180--104204},
 publisher = {Curran Associates, Inc.},
 title = {Optimal Flow Matching: Learning Straight Trajectories in Just One Step},
 url = {https://proceedings.neurips.cc/paper_files/paper/2024/file/bc8f76d9caadd48f77025b1c889d2e2d-Paper-Conference.pdf},
 volume = {37},
 year = {2024}
}

@article{meanflow,
  title={Mean Flows for One-step Generative Modeling},
  author={Zhengyang Geng and Mingyang Deng and Xingjian Bai and J. Zico Kolter and Kaiming He},
  journal={arXiv preprint arXiv:2505.13447},
  year={2025},
  url={https://arxiv.org/abs/2505.13447}
}

@inproceedings{ICNN2017,
  title = 	 {Input Convex Neural Networks},
  author =       {Brandon Amos and Lei Xu and J. Zico Kolter},
  booktitle = 	 {Proceedings of the 34th International Conference on Machine Learning},
  pages = 	 {146--155},
  year = 	 {2017},
  volume = 	 {70},
  series = 	 {Proceedings of Machine Learning Research},
  month = 	 {06--11 Aug},
  publisher =    {PMLR},
  url = 	 {https://proceedings.mlr.press/v70/amos17b.html}
}

@article{yang2023diffusion,
  title={Diffusion models: A comprehensive survey of methods and applications},
  author={Yang, Ling and Zhang, Zhilong and Song, Yang and Hong, Shenda and Xu, Runsheng and Zhao, Yue and Zhang, Wentao and Cui, Bin and Yang, Ming-Hsuan},
  journal={ACM Computing Surveys},
  volume={56},
  number={4},
  pages={1--39},
  year={2023},
  publisher={ACM New York, NY, USA}
}

@article{liu2022rectified,
  title={Rectified Flow: A Marginal Preserving Approach to Optimal Transport},
  author={Qiang Liu},
  journal={arXiv preprint arXiv:2209.14577},
  year={2022},
  url={https://arxiv.org/abs/2209.14577}
}

@inproceedings{korotin2021bm,
 author = {Korotin, Alexander and Li, Lingxiao and Genevay, Aude and Solomon, Justin M and Filippov, Alexander and Burnaev, Evgeny},
 booktitle = {Advances in Neural Information Processing Systems},
 editor = {M. Ranzato and A. Beygelzimer and Y. Dauphin and P.S. Liang and J. Wortman Vaughan},
 pages = {14593--14605},
 publisher = {Curran Associates, Inc.},
 title = {Do Neural Optimal Transport Solvers Work? A Continuous Wasserstein-2 Benchmark},
 url = {https://proceedings.neurips.cc/paper_files/paper/2021/file/7a6a6127ff85640ec69691fb0f7cb1a2-Paper.pdf},
 volume = {34},
 year = {2021}
}

@article{HJPINN,
title = {Physics-informed neural networks: A deep learning framework for solving forward and inverse problems involving nonlinear partial differential equations},
journal = {Journal of Computational Physics},
volume = {378},
pages = {686-707},
year = {2019},
issn = {0021-9991},
doi = {https://doi.org/10.1016/j.jcp.2018.10.045},
url = {https://www.sciencedirect.com/science/article/pii/S0021999118307125},
author = {M. Raissi and P. Perdikaris and G.E. Karniadakis}
}

@inproceedings{rezende2015variational,
  title = 	 {Variational Inference with Normalizing Flows},
  author = 	 {Rezende, Danilo and Mohamed, Shakir},
  booktitle = 	 {Proceedings of the 32nd International Conference on Machine Learning},
  pages = 	 {1530--1538},
  year = 	 {2015},
  volume = 	 {37},
  series = 	 {Proceedings of Machine Learning Research},
  publisher =    {PMLR},
  url = 	 {https://proceedings.mlr.press/v37/rezende15.html},
}

@article{papamakarios2021normalizing,
  title={Normalizing flows for probabilistic modeling and inference},
  author={Papamakarios, George and Nalisnick, Eric and Rezende, Danilo Jimenez and Mohamed, Shakir and Lakshminarayanan, Balaji},
  journal={Journal of Machine Learning Research},
  volume={22},
  number={57},
  pages={1--64},
  year={2021}
}

@article{fan2023neural,
  title={Neural monge map estimation and its applications},
  author={Fan, Jiaojiao and Liu, Shu and Ma, Shaojun and Zhou, Hao-Min and Chen, Yongxin},
  journal={Transactions on Machine Learning Research},
  year={2023}
}

@article{li2025dpot,
  title={DPOT: A DeepParticle Method for Computation of Optimal Transport with Convergence Guarantee},
  author={Yingyuan Li and Aokun Wang and Zhongjian Wang},
  journal={arXiv preprint arXiv:2506.23429},
  year={2025},
  url={https://arxiv.org/abs/2506.23429}
}

@article{wang2022deepparticle,
  title={DeepParticle: learning invariant measure by a deep neural network minimizing Wasserstein distance on data generated from an interacting particle method},
  author={Wang, Zhongjian and Xin, Jack and Zhang, Zhiwen},
  journal={Journal of Computational Physics},
  volume={464},
  pages={111309},
  year={2022},
  url = {https://www.sciencedirect.com/science/article/pii/S0021999122003710},
  publisher={Elsevier}
}

@inproceedings{pidhorskyi2020adversarial,
  title={Adversarial latent autoencoders},
  author={Pidhorskyi, Stanislav and Adjeroh, Donald A and Doretto, Gianfranco},
  booktitle={Proceedings of the IEEE/CVF conference on computer vision and pattern recognition},
  pages={14104--14113},
  year={2020}
}

@article{ROY2023472,
 title={Deep learning-accelerated computational framework based on physics informed neural network for the solution of linear elasticity},
  author={Roy, Arunabha M and Bose, Rikhi and Sundararaghavan, Veera and Arr{\'o}yave, Raymundo},
  journal={Neural Networks},
  volume={162},
  pages={472--489},
  year={2023},
  publisher={Elsevier}
}

@article{ROY2023106049,
title = {A data-driven physics-constrained deep learning computational framework for solving von Mises plasticity},
author={Roy, Arunabha M and Guha, Suman},
  journal={Engineering Applications of Artificial Intelligence},
  volume={122},
  pages={106049},
  year={2023},
  publisher={Elsevier}
}

@article{BOSE2024107483,
  title={Invariance embedded physics-infused deep neural network-based sub-grid scale models for turbulent flows},
  author={Bose, Rikhi and Roy, Arunabha M},
  journal={Engineering Applications of Artificial Intelligence},
  volume={128},
  pages={107483},
  year={2024},
  publisher={Elsevier}
}

@article{ROY2024105570,
  title={Physics-infused deep neural network for solution of non-associative Drucker--Prager elastoplastic constitutive model},
  author={Roy, Arunabha M and Guha, Suman and Sundararaghavan, Veera and Arr{\'o}yave, Raymundo},
  journal={Journal of the Mechanics and Physics of Solids},
  volume={185},
  pages={105570},
  year={2024},
  publisher={Elsevier}
}

@article{tong2023conditional,
  title={Conditional flow matching: Simulation-free dynamic optimal transport},
  author={Tong, Alexander and Malkin, Nikolay and Huguet, Guillaume and Zhang, Yanlei and Rector-Brooks, Jarrid and Fatras, Kilian and Wolf, Guy and Bengio, Yoshua},
  journal={arXiv preprint arXiv:2302.00482},
  volume={2},
  number={3},
  year={2023}
}

@inproceedings{
liu2023flow,
title={Flow Straight and Fast: Learning to Generate and Transfer Data with Rectified Flow},
author={Xingchao Liu and Chengyue Gong and Qiang Liu},
booktitle={International Conference on Learning Representations (ICLR)},
year={2023},
url={https://openreview.net/forum?id=XVjTT1nw5z}
}

@article{sirignano2018dgm,
  title={DGM: A deep learning algorithm for solving partial differential equations},
  author={Sirignano, Justin and Spiliopoulos, Konstantinos},
  journal={Journal of Computational Physics},
  volume={375},
  pages={1339--1364},
  year={2018},
  publisher={Elsevier}
}

@article{han2018solving,
  title={Solving high-dimensional partial differential equations using deep learning},
  author={Han, Jiequn and Jentzen, Arnulf and E, Weinan},
  journal={Proceedings of the National Academy of Sciences},
  volume={115},
  number={34},
  pages={8505--8510},
  year={2018},
  publisher={National Academy of Sciences}
}

@article{e2018deep,
  title={The deep Ritz method: a deep learning-based numerical algorithm for solving variational problems},
  author={Yu, Bing and others},
  journal={Communications in Mathematics and Statistics},
  volume={6},
  number={1},
  pages={1--12},
  year={2018},
  publisher={Springer}
}

@article{roy2024combining,
  title={Combining crystal plasticity and phase field model for predicting texture evolution and the influence of nuclei clustering on recrystallization path kinetics in Ti-alloys},
  author={Roy, Arunabha M and Ganesan, Sriram and Acar, Pinar and Arr{\'o}yave, Raymundo and Sundararaghavan, V},
  journal={Acta Materialia},
  volume={266},
  pages={119645},
  year={2024},
  publisher={Elsevier}
}

@article{mccann1997convexity,
  title={A convexity principle for interacting gases},
  author={McCann, Robert J},
  journal={Advances in mathematics},
  volume={128},
  number={1},
  pages={153--179},
  year={1997},
  publisher={Elsevier}
}

@inproceedings{de2021diffusion,
  title={Diffusion schr{\"o}dinger bridge with applications to score-based generative modeling},
  author={De Bortoli, Valentin and Thornton, James and Heng, Jeremy and Doucet, Arnaud},
  booktitle={Advances in Neural Information Processing Systems},
  volume={34},
  pages={17695--17709},
  year={2021}
}

@inproceedings{esser2024scaling,
  title={Scaling rectified flow transformers for high-resolution image synthesis},
  author={Esser, Patrick and Kulal, Sumith and Blattmann, Andreas and Entezari, Rahim and M{\"u}ller, Jonas and Saini, Harry and Levi, Yam and Lorenz, Dominik and Sauer, Axel and Boesel, Frederic and others},
  booktitle={Forty-first international conference on machine learning},
  year={2024}
}

@article{chen2016relation,
  title={On the relation between optimal transport and Schr{\"o}dinger bridges: A stochastic control viewpoint},
  author={Chen, Yongxin and Georgiou, Tryphon T and Pavon, Michele},
  journal={Journal of Optimization Theory and Applications},
  volume={169},
  number={2},
  pages={671--691},
  year={2016},
  publisher={Springer}
}

@inproceedings{rombach2022high,
  title={High-resolution image synthesis with latent diffusion models},
  author={Rombach, Robin and Blattmann, Andreas and Lorenz, Dominik and Esser, Patrick and Ommer, Bj{\"o}rn},
  booktitle={Proceedings of the IEEE/CVF conference on computer vision and pattern recognition},
  pages={10684--10695},
  year={2022}
}

@article{raissi2019physics,
  title={Physics-informed neural networks: A deep learning framework for solving forward and inverse problems involving nonlinear partial differential equations},
  author={Raissi, Maziar and Perdikaris, Paris and Karniadakis, George E},
  journal={Journal of Computational Physics},
  volume={378},
  pages={686--707},
  year={2019},
  publisher={Elsevier}
}

@article{nakamura2021adaptive,
  title={Adaptive Deep Learning for High-Dimensional Hamilton--Jacobi--Bellman Equations},
  author={Nakamura-Zimmerer, Tenavi and Gong, Qi and Kang, Wei},
  journal={SIAM Journal on Scientific Computing},
  volume={43},
  number={2},
  pages={A1221--A1247},
  year={2021},
  publisher={SIAM}
}

\end{document}